\documentclass[10pt,twocolumn,letterpaper]{article}

\usepackage{array}
\usepackage[pagenumbers]{wacv} % To force page numbers, e.g. for an arXiv version

\usepackage[normalem]{ulem}

\definecolor{wacvblue}{rgb}{0.21,0.49,0.74}
\usepackage[pagebackref,breaklinks,colorlinks,allcolors=wacvblue]{hyperref}

\def\wacvPaperID{1097} % *** Enter the WACV Paper ID here
\def\confName{WACV}
\def\confYear{2027}

\title{Angular Supervision for Visual Object Grounding \\ in Autoregressive Vision-Language Models}

\author{Parsa Esmaeilkhani\\
Temple University\\
Philadelphia, USA\\
{\tt\small parsa.esmaeilkhani@temple.edu}
\and
Longin Jan Latecki\\
Temple University\\
Philadelphia, USA\\
{\tt\small latecki@temple.edu}
}

\begin{document}
\maketitle

\begin{abstract}

Autoregressive vision-language models (VLMs) struggle to ground generated text in the correct image regions, and existing explainability methods, such as attention maps and Logit Lens, often produce poor spatial explanations due to weak text-to-vision attention and sink tokens. We propose \textbf{Angular Relevance} ($R_L$), a simple visualization measure based on cosine similarity between visual token embeddings and unembedding-matrix rows. Because $R_L$ is invariant to token norm, it produces substantially sharper and more faithful heatmaps than attention maps or Logit Lens on out-of-the-box models. Without any training, $R_L$ yields competitive referring-expression segmentation (RES), outperforming prior diagnostic maps and CLIP-based training-free methods. To preserve localized object information during cross-modal mixing, we introduce \textbf{Angular Object-Aware Loss (AOL)}, an auxiliary loss that directly supervises visual token embeddings to align angularly with the vocabulary directions of their corresponding object concepts. Finetuning with AOL alongside the standard next-token prediction objective significantly sharpens object localization in heatmaps, showing that the VLM's own visual tokens can be used directly for grounding and improving RES, without training any segmentation heads and without degrading general VQA performance. We further demonstrate that the learned grounding also transfers to category-level segmentation of object categories not seen during training. Code: \href{https://github.com/ParsaKhani/Angular_object_grounding}{GitHub}.
\end{abstract}

% \vspace{-8mm}
\section{Introduction}

Autoregressive vision--language models (VLMs) have become increasingly capable at generating textual answers conditioned on images \cite{liu2023visual, bai2025qwen2, alayrac2022flamingo}. In these models, the image is represented as a sequence of image patch features projected into the LLM embedding space; we refer to these projected image tokens as \emph{visual tokens}. These visual tokens are concatenated with text tokens and processed jointly by the LLM through self-attention. Despite this joint processing, it remains difficult to determine whether a generated word is grounded in the correct visual evidence. This issue is especially important for object-level grounding tasks, where the model should not only generate the correct concept, but also localize the corresponding visual evidence in the image \cite{yu2016modeling,rasheed2024glamm,lai2024lisa}.

\begin{figure*}[hbtp]
    \centering
    \includegraphics[width=0.9\textwidth]{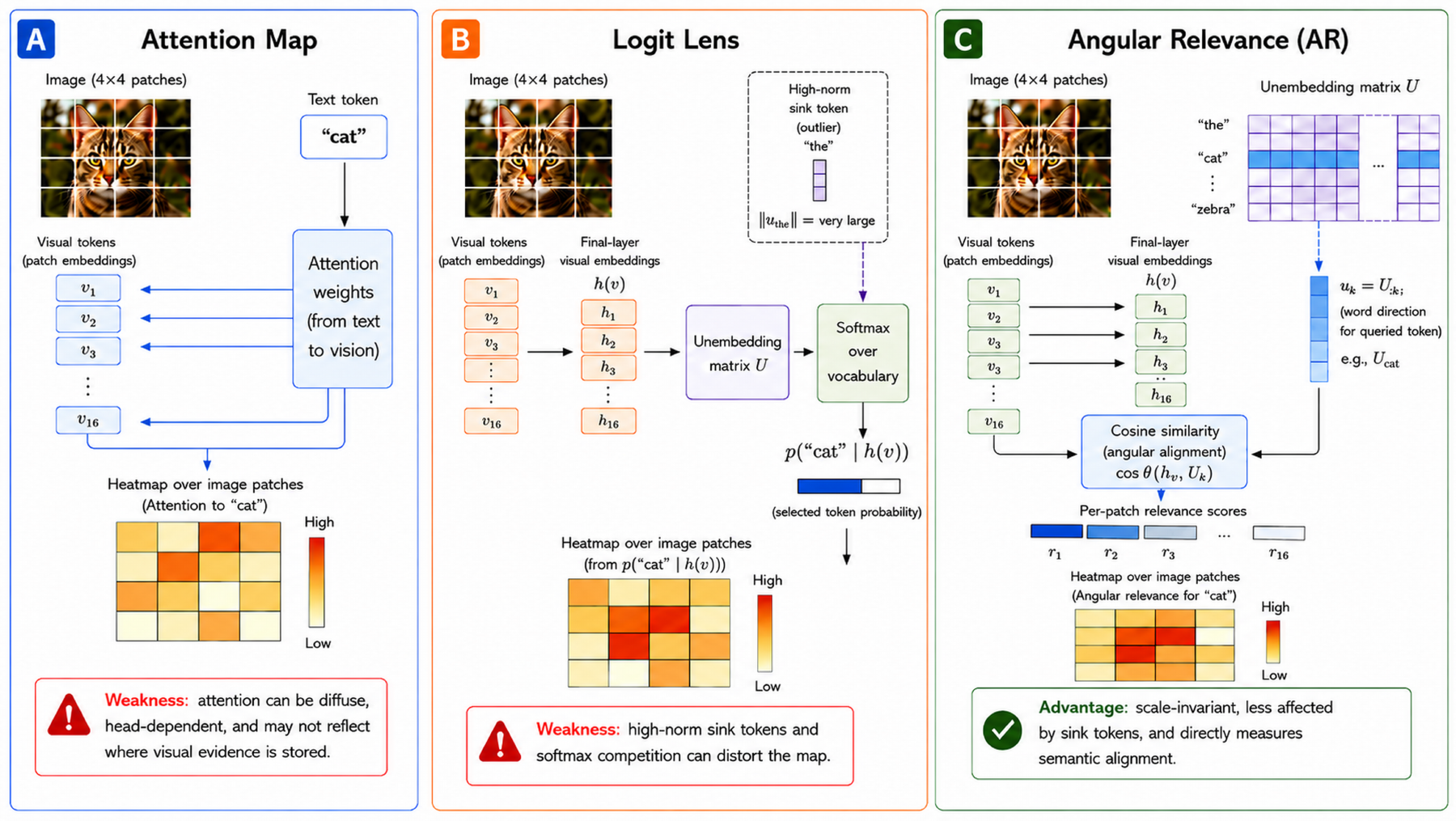}
    % {figs/interpretability_tools_font_larger_paper_4x_1200dpi.pdf}    
    \caption{
        Mechanistic comparison of three diagnostic maps for autoregressive vision-language models. 
        Attention maps visualize text-to-vision attention weights, Logit Lens maps visual embeddings through the unembedding matrix and softmax to obtain queried-token probabilities, while Angular Relevance directly measures cosine alignment between final-layer visual embeddings and the queried vocabulary direction (rows of the unembedding matrix).
    }
    \label{fig:interpretation_methods}
    \vspace{-5mm}
\end{figure*}

Existing diagnostic tools provide partial views into this process, but they are often disappointing as spatial explanations: attention maps indicate where text tokens attend, while Logit Lens estimates what vocabulary items a visual token may predict. Neither directly measures whether a visual token is semantically aligned with a queried object concept, e.g., in an image containing a cat, whether the visual tokens corresponding to image patches covering the cat are aligned with the vocabulary direction of ``cat''. Earlier explainability methods in computer vision, such as CAM and Grad-CAM, produced class-specific heatmaps by projecting convolutional evidence back to image space \cite{zhou2016cam,selvaraju2017gradcam}. With Vision Transformers and CLIP-like models, images became sequences of visual tokens, and attribution methods increasingly relied on attention or gradient signals over token interactions \cite{dosovitskiy2021vit,radford2021learning,bousselham2024legrad}. 

In autoregressive VLMs, however, the explanation target is no longer a fixed class logit, but a generated text token and its evolving distribution over the vocabulary. This makes the problem different: we need to understand how individual visual tokens support a particular generated word.
A common diagnostic is to visualize attention weights from generated text tokens to visual tokens, providing an intuitive link between generated words and image regions. However, attention is an indirect and often unstable explanation: it can be diffuse, strongly layer- and head-dependent, absorbed by sink tokens, or misplaced over irrelevant regions rather than where object-level evidence is stored. 
% Recent studies show that visual grounding can rely on only a few specialized localization heads, requiring careful criteria to select vision-centric heads \cite{kang2025your}, while spatial reasoning and grounding failures are associated with weak or misdirected attention over relevant regions \cite{chen2025spatialreasoninghardvlms,esmaeilkhani2025directvisualgroundingdirecting}.
Recent studies link visual grounding to a few specialized localization heads~\cite{kang2025your} and show that grounding failures often arise from weak or misdirected attention~\cite{chen2025spatialreasoninghardvlms,esmaeilkhani2025directvisualgroundingdirecting}.

A more natural diagnostic for this setting is Logit Lens, which projects hidden states through the final unembedding layer to inspect which text tokens the model is leaning toward \cite{logitlens2020,LogitLensICLR2025}. For a visual token $v$ with hidden representation $h_l(v)$ at layer $l$, Logit Lens visualizes the probability of a queried token $t$, written as $p_{\theta}(t \mid h_l(v))$. However, because this probability is obtained after the unembedding projection and softmax, the resulting heatmaps can be distorted by vocabulary competition and high-norm visual tokens \cite{xiao2024efficientstreaminglanguagemodels}. As we demonstrate, this makes Logit Lens ineffective at visualizing the semantic content of image patches in modern VLMs such as Qwen2.5-VL: high-norm patches can dominate the map, suppress important object details, and make it difficult to distinguish foreground from background. 
To address these challenges, we propose \textbf{Angular Relevance}, denoted by $R_L$, a simple visualization method that focuses on \textbf{angular relationships rather than magnitudes} between deep embeddings of visual tokens and selected rows of the VLM unembedding matrix at layer $L$. Since $R_L$ is based on cosine similarity, it is invariant to rescaling of visual-token embeddings, i.e., $R_L(t,\gamma h_L(v))=R_L(t,h_L(v))$ for any $\gamma>0$. This makes the relevance score depend on semantic alignment between the visual embedding and the queried text direction, rather than on embedding norm. Fig.~\ref{fig:interpretation_methods} illustrates this distinction.

Although Angular Relevance produces substantially cleaner heatmaps than attention maps and Logit Lens, it remains a diagnostic tool: it reveals how object information is represented, but it does not by itself improve the model's internal grounding. In autoregressive VLMs, visual-token embeddings are repeatedly mixed with text-token embeddings through self-attention, and localized object information can become diffused across the sequence. This motivates the central question of this work:
\textbf{Can we preserve localized object information in the model's own visual-token representations without sacrificing its general language and instruction-following ability?} 

To answer this question, we introduce \textbf{Angular Object-Aware Loss} (AOL), an auxiliary objective that directly supervises visual-token embeddings in the same unembedding space used by Angular Relevance $R_L$. Given a queried object, AOL encourages the embeddings of visual tokens overlapping the object region to align with the corresponding vocabulary direction, while discouraging background tokens from aligning with it. For instance, for the query ``cat'', AOL aligns visual-token embeddings from cat patches with the ``cat'' vocabulary direction, while pushing non-cat patches away.
% In this way, AOL turns the diagnostic signal of Angular Relevance into a representation-learning objective: rather than merely observing whether visual tokens are object-aligned, we directly improve their object-level alignment inside the VLM's own embedding space. 
In this way, AOL turns the diagnostic signal of Angular Relevance into a representation-learning objective, directly improving object-level alignment inside the VLM's own embedding space.
% When utilized together with the next-token prediction (NTP) loss, AOL constrains the mixing of visual and text tokens by preserving image content in visual tokens representing local image patches. This enables direct localization of text tokens in the image and improves the spatial object-awareness of internal visual embeddings, leading to stronger performance on referring expression (RES) without degrading the model's general language and instruction-following ability.
When combined with the next-token prediction (NTP) loss, AOL preserves local image content in visual tokens during cross-modal mixing, enabling direct localization of text tokens and improving RES performance without degrading the model's language and instruction-following ability.
We refer to Fig.~\ref{fig:qual_3x5_square} for visual comparison.

This design differs from segmentation-oriented VLMs that improve grounding by attaching a mask decoder or dense prediction module and supervising the resulting architecture for mask prediction \cite{lai2024lisa,rasheed2024glamm}. Such dense prediction pathways can introduce tension with preserving the language ability of the original autoregressive model \cite{wu2025f}. In contrast, AOL uses object masks only to provide \textbf{per-token supervision for the model's existing visual embeddings}, rather than as targets for training a separate mask-prediction head.
% Even though the visual explainability of $R_L$ provides a significant improvement over Logit Lens, it is still far from being perfect.
% We attribute this to unconstrained mixing of visual tokens with text tokens in the self-attention layers, which diffuses the localized information from visual tokens to the other tokens.
% Hence, the localized content of visual tokens is mostly lost. This yields the following question:
% \textbf{Can we preserve the localized information of visual tokens in the mixing process without reducing the performance of a VLM on downstream tasks?}
% To answer this question, we propose to add a new loss, Object Awareness Loss (AOL), in order to align the visualization ability of VLMs with generated content.
% In other words, AOL grounds the generated content in the image tokens.
% When used together with the next-token prediction (NTP) loss, AOL constraints mixing of visual and text tokens by grounding the image content in visual tokens representing local image patches,
% e.g., text token "cat" patches representing parts of the cat depicted in the image. 
% This makes a direct localization of text tokens in images possible without attaching specialized (e.g., segmentation),
% as we demonstrate experimentally.
Our main contributions can be summarized as:
\begin{itemize}
    \item We propose a simple Angular Relevance measure $R_{L}$ of image patch tokens wrt text tokens.
    We demonstrate experimentally that $R$ provides a stronger explainability signal than Logit Lens and attention visualization on out-of-the-box models.
    \item We propose an auxiliary loss, called Angular object-aware loss (AOL), that forces visual tokens to preserve their localized information about the image content.
    We show that AOL provides a more direct and stronger gradient signal to visual tokens than NTP loss.
    Moreover, AOL significantly reduces high norm tokens among visual tokens, which are a byproduct of softmax used in NTP. 
    \item We obtain a substantial improvement in the quality of heatmaps when $R_L$ is applied after finetuning with AOL + NTP, so that the heatmaps can be used to obtain strong RES results. 
    This implies that visual tokens can be directly utilized for the visual grounding of text tokens.
    %without introducing a trainable segmentation decoder or segmentation head into the VLM. 
    This shows that AOL keeps visual information localized and does not degrade the language capabilities of the VLM on downstream tasks.
    \item We demonstrate experimentally that the visual grounding ability generalizes to object classes that exhibit no corresponding category in the training data.

\end{itemize}

%%%%%%%%%%%%%%%%%%%%%%%

% \vspace{-3mm}
\section{Preliminaries}
% \vspace{-2mm}
We focus on autoregressive VLMs,
like LLava-v1.5 \cite{liu2023visual} and Qwen2.5-VL \cite{bai2025qwen2}.
The input image (or images) is first passed through a frozen visual encoder (typically a ViT, CLIP, or DINO) to extract image tokens and map them into a language embedding space with a linear projection module.
Subsequently, the mapped image tokens, often called visual tokens, are used as part of the LLM input, 
together with the text input (and special tokens). 
% As visual tokens are transformed through the layers of an LLM, their embeddings change.
% In particular, the embeddings of visual tokens are mixed with the embeddings of language tokens in the attention layers of the LLM.
% In this process, the localized information carried by visual tokens often becomes diffused among the predominantly textual tokens as they propagate through the LLM layers.
% In particular, the information of visual tokens may be transferred to language tokens,
% as nicely demonstrated in \cite{whatsimagedeepdCVPR2025}.
% In an extreme case, the visual tokens are ignored and the LLM can even hallucinate an image description for an empty image, as demonstrated in  \cite{payattentionimageECCV2024,tong2024cambrian}.
% Another piece of evidence that visual tokens are often overlooked among all LLM tokens is that after removing half of the visual tokens, the VLM performance does not decrease \cite{chen2024imageworth12tokens}. 
% A consequence is that the connection of visual tokens to the local content of corresponding image regions is lost in this process.

As visual tokens pass through the layers of an LLM, their embeddings are mixed with language tokens in the attention layers, causing localized visual information to diffuse among textual tokens. This can lead to visual information being transferred to language tokens \cite{whatsimagedeepdCVPR2025}, or even ignored entirely, with the LLM generating descriptions for empty images \cite{payattentionimageECCV2024}. Additionally, removing half of the visual tokens does not reduce VLM performance, suggesting that the connection between visual tokens and their corresponding image regions is often lost \cite{chen2024imageworth12tokens}.

\vspace{-2mm}

\subsection{Core Idea of Logit Lens}
\vspace{-2mm}

While Logit Lens was originally proposed for visualization in LLMs, it can be naturally applied to autoregressive VLMs.
% {While Logit Lens was originally proposed for LLMs, prior VLM interpretability work has applied the LM unembedding matrix to visual-token representations across layers~\cite{neo2025interpretingvisualinformationprocessing}, providing empirical precedent for interpreting visual representations in the vocabulary space.}

In an LLM, an unembedding matrix $U$ maps the last latent representation $h_L(t_{n-1})$ of a text token $t_{n-1}$ to a logit vector over the vocabulary tokens.
After applying the softmax to the logit vector, the logits are converted to a probability distribution over the vocabulary $p_{\theta}(t \mid h_l(v))$ to select the next token $t_n$, where $t$ varies over all vocabulary tokens, and $\theta$ represents the network parameters.
Instead of applying the unembedding matrix $U$ to $h_L(t_{n-1})$, we can apply it to a deep embedding of visual tokens $h_l(v)$, where $l$ is one of the VLM layers $1:L$. 
For a given visual token $v$, we define the relevance of text token $t$ to $v$ as 
\begin{equation} \label{eq:LogitLens}
    \text{LogitLens}_l(t, v) = p_{\theta}(t \mid h_l(v))
\end{equation}

% \vspace{-3mm}

\section{Methodology}

% \vspace{-1mm}

\subsection{Relevance Measure for Explainability}
\vspace{-2mm}

% We propose to define the relevance score of a text token $t$ to a visual token $v$ as the cosine similarity:
% \begin{equation}
% R_L(t,v) = \cos(U_t^\top, h_L(v)) =
% \frac{U_t^\top h_L(v)}{\|U_t^\top\|\| h_L(v)\|},
% \label{eq:relevance}
% \end{equation}
% where $h_L(v)$ is a deep embedding vector from the last layer $L$ and $U_t^\top$ is a row of the VLM unembedding matrix representing text token $t$.

We propose to define the relevance score of a text token $t$ to a visual token $v$ as the cosine similarity:
\begin{equation}
R_L(t,v) = \cos(U_t, h_L(v)) =
\frac{U_t\, h_L(v)}{\|U_t\|\,\| h_L(v)\|},
\label{eq:relevance}
\end{equation}
where $U \in \mathbb{R}^{|V| \times d}$ is the VLM unembedding matrix mapping a $d$-dimensional hidden representation to a logit vector over the vocabulary $V$, $h_L(v) \in \mathbb{R}^d$ is a deep embedding vector from the last layer $L$, and $U_t \in \mathbb{R}^{1\times d}$ denotes the $t$-th row of $U$, treated as a $d$-dimensional vector so that $U_t\, h_L(v)$ denotes their inner product.

We use the final layer since it is the only layer directly aligned with the
unembedding matrix for vocabulary prediction.
% Also, If a given referring expression $\tau$ is represented by more than one text token, 
% % we take an average of their corresponding rows $U_t^\top$.
% we first group tokens belonging
% to the same semantic concept (e.g., nouns and descriptors), average the
% corresponding rows of the unembedding matrix within each concept, and then
% average across concepts to obtain a single query direction $U_t$. For example, in ``the lady with the blue shirt'', ``lady'' and ``shirt'' are nouns
% and ``blue'' is a descriptor of ``shirt''; we average the token embeddings within
% each word (if split into subwords) and then average across the three concepts to
% form the final query direction.
If a referring expression $\tau$ contains multiple content words, we first construct a direction for each content word by averaging the corresponding rows of the unembedding matrix when the word is split into multiple subword tokens. We then average these word-level directions to obtain a single query direction $U_t$. For example, for ``the lady with the blue shirt'', we construct directions for ``lady'', ``blue'', and ``shirt'', and then average these three directions to obtain the final query direction.

A heat map visualizing the VLM's internal confidence of an object described with a text token $t$ is given by 
% $\{ R_L(t,v) | v \in V \}$ reshaped and unscaled to the size of the input image,
% where $V$ is the set of the image patch tokens.
$\{ R_L(t,v) \mid v \in V \}$, 
where $V$ is the set of the image patch tokens. We place the patch scores back into their original spatial positions to form a 2D grid, and resize this grid to match image resolution.
We then apply $\mathrm{ReLU}$ to suppress negative cosine similarities (anti-aligned directions) and apply light Gaussian smoothing to reduce noise and fill small gaps.

\begin{figure*}[hbtp]
\centering
\setlength{\tabcolsep}{2pt}

\begin{tabular}{ccccc}

% ================= Row 1 =================
\begin{minipage}[t]{0.19\textwidth}\centering
\includegraphics[width=\linewidth,height=\linewidth]{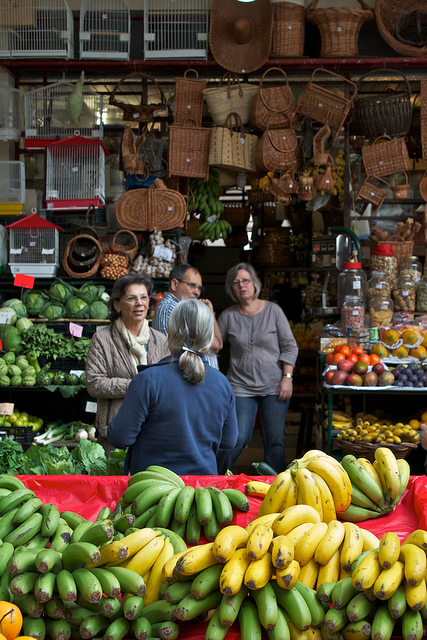}\par\vspace{-2pt}
{\footnotesize Lady in a blue shirt}\par\vspace{1pt}
\end{minipage} &
\begin{minipage}[t]{0.19\textwidth}\centering
\includegraphics[width=\linewidth,height=\linewidth]{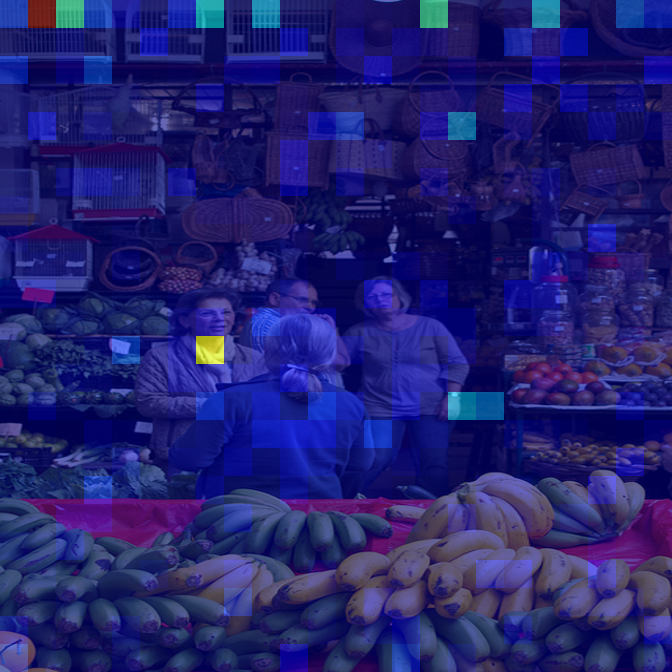}
\end{minipage} &
\begin{minipage}[t]{0.19\textwidth}\centering
\includegraphics[width=\linewidth,height=\linewidth]{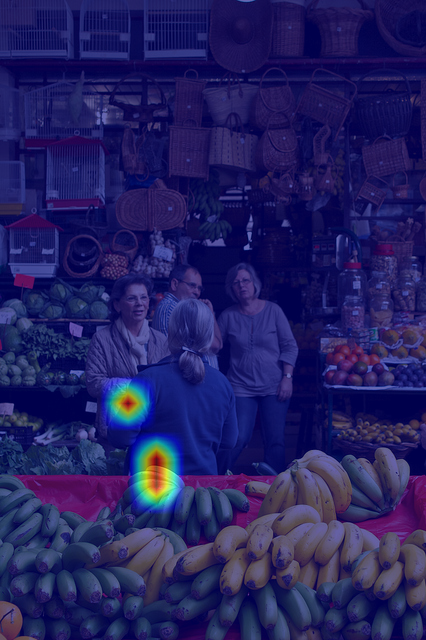}
\end{minipage} &
\begin{minipage}[t]{0.19\textwidth}\centering
\includegraphics[width=\linewidth,height=\linewidth]{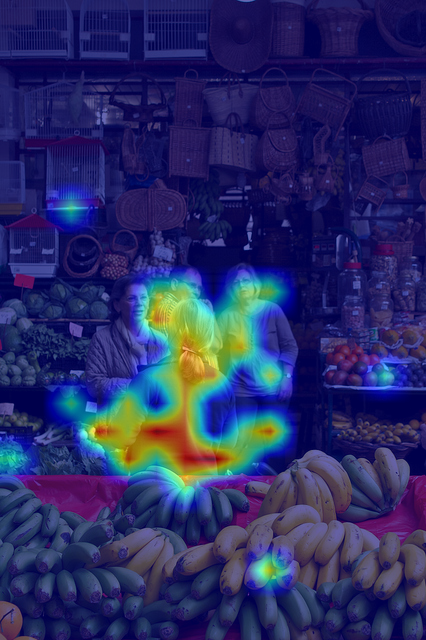}
\end{minipage} &
\begin{minipage}[t]{0.19\textwidth}\centering
\includegraphics[width=\linewidth,height=\linewidth]{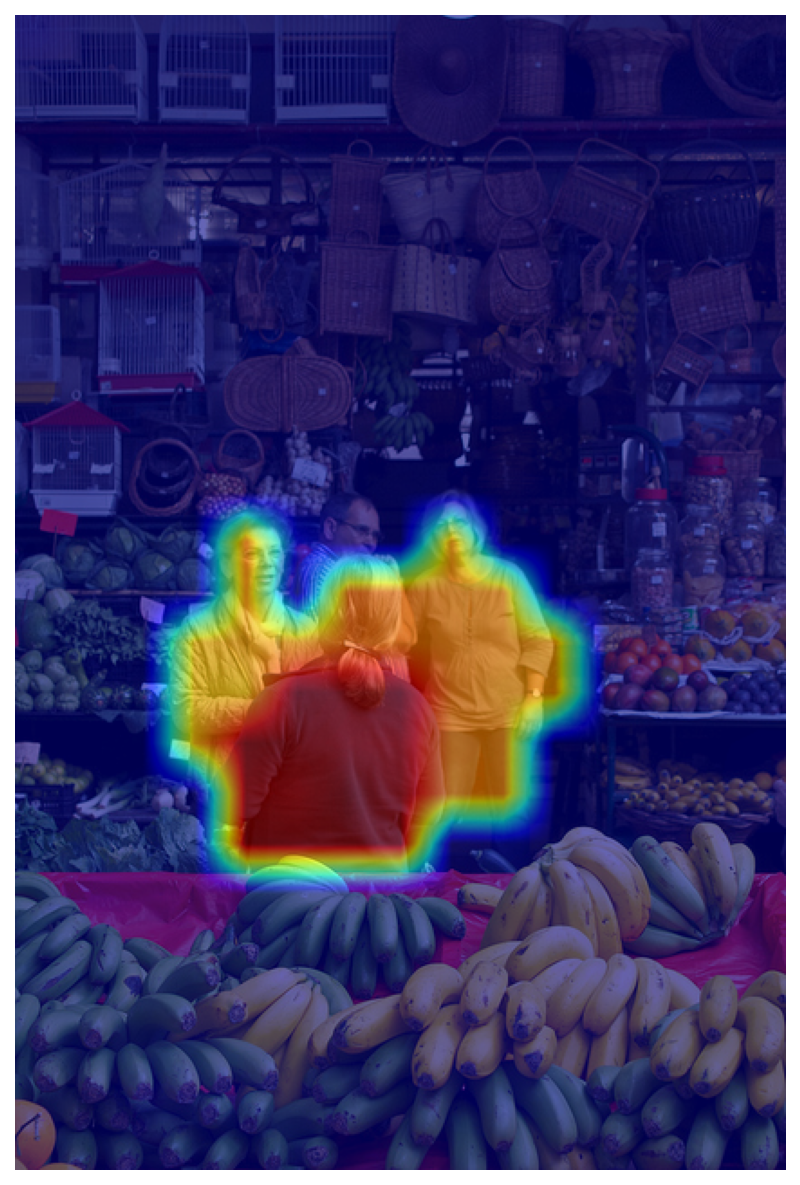}
\end{minipage} \\[2pt]

% % ================= Row 2 =================
\begin{minipage}[t]{0.19\textwidth}\centering
\includegraphics[width=\linewidth,height=\linewidth]{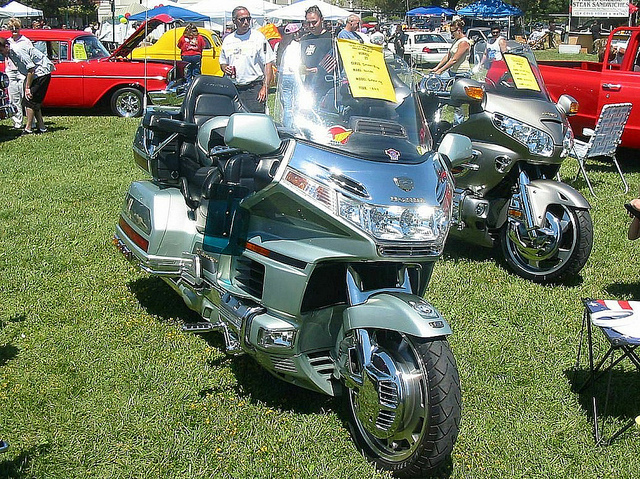}\par\vspace{-2pt}
{\footnotesize Motorcycle on the right}\par\vspace{1pt}
\end{minipage} &
\begin{minipage}[t]{0.19\textwidth}\centering
\includegraphics[width=\linewidth,height=\linewidth]{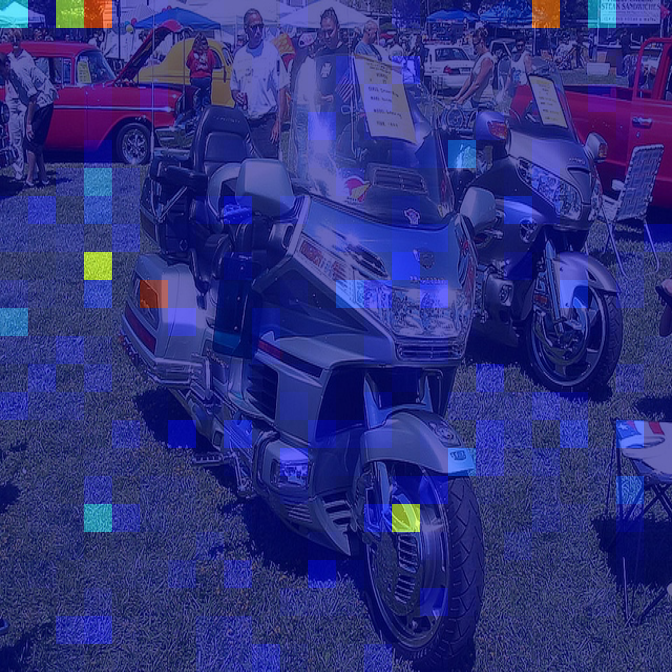}
\end{minipage} &
\begin{minipage}[t]{0.19\textwidth}\centering
\includegraphics[width=\linewidth,height=\linewidth]{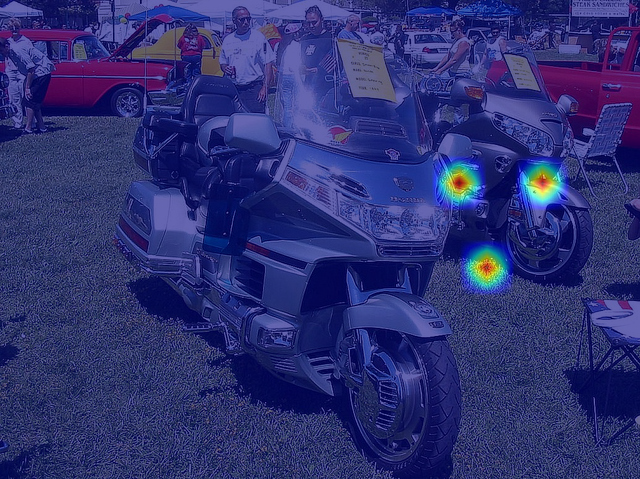}
\end{minipage} &
\begin{minipage}[t]{0.19\textwidth}\centering
\includegraphics[width=\linewidth,height=\linewidth]{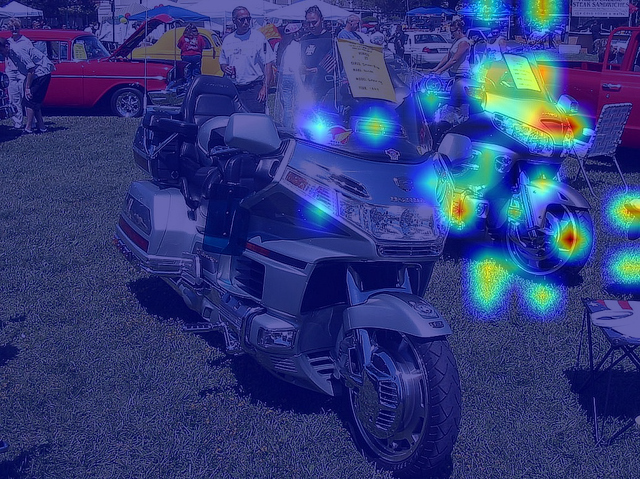}
\end{minipage} &
\begin{minipage}[t]{0.19\textwidth}\centering
\includegraphics[width=\linewidth,height=\linewidth]{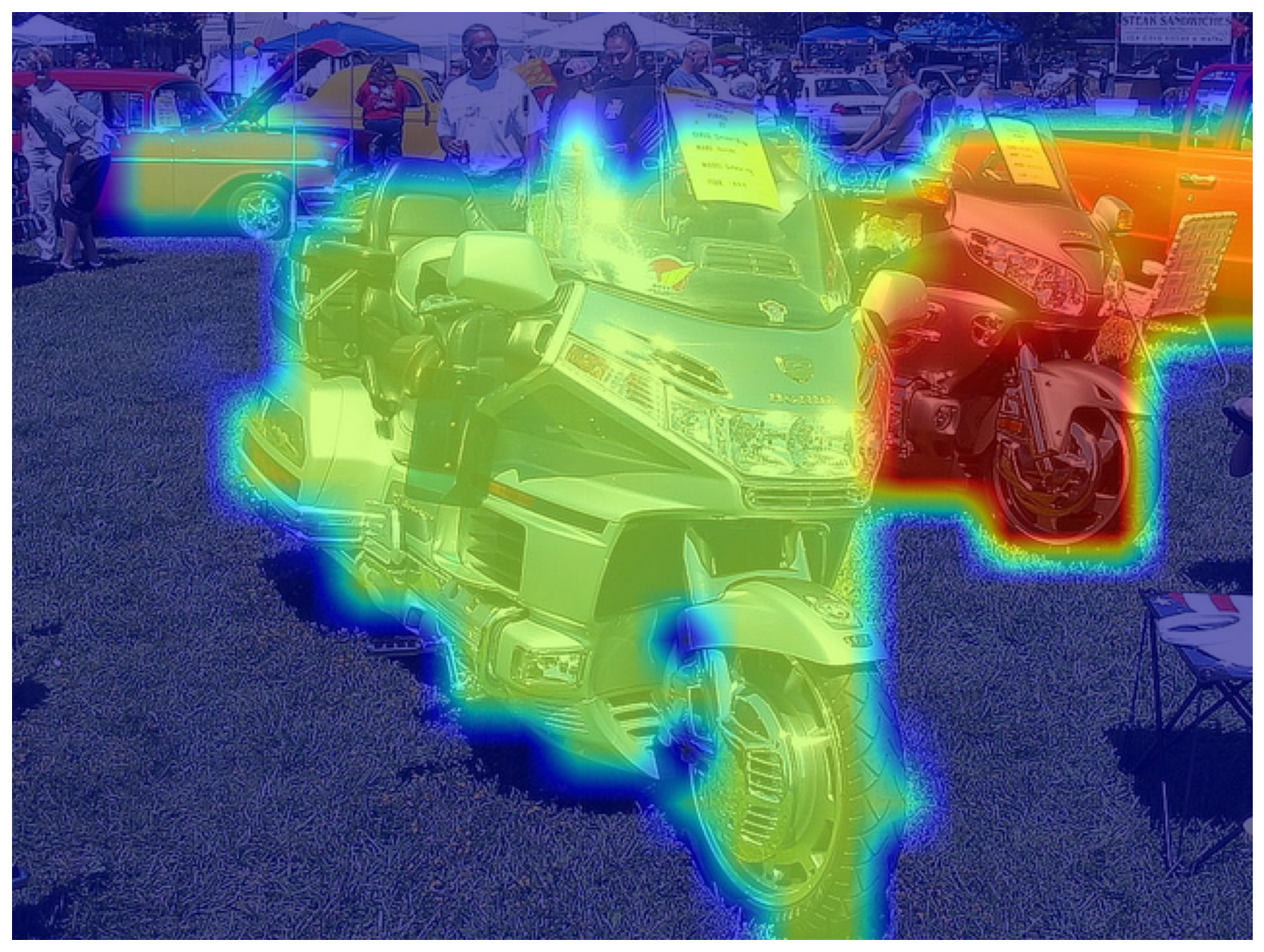}
\end{minipage} \\[2pt]

% ================= Row 3 =================
\begin{minipage}[t]{0.19\textwidth}\centering
\includegraphics[width=\linewidth,height=\linewidth]{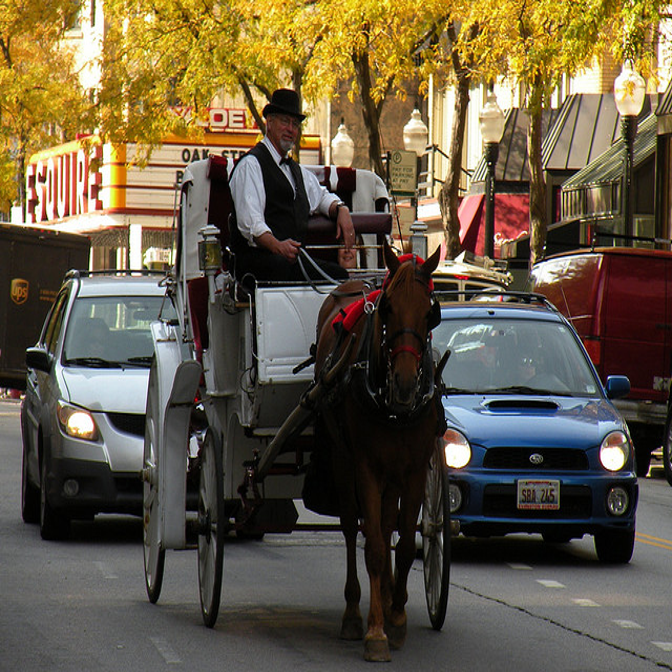}\par\vspace{-2pt}
{\footnotesize Far left gray car}\par\vspace{1pt}
\end{minipage} &
\begin{minipage}[t]{0.19\textwidth}\centering
\includegraphics[width=\linewidth,height=\linewidth]{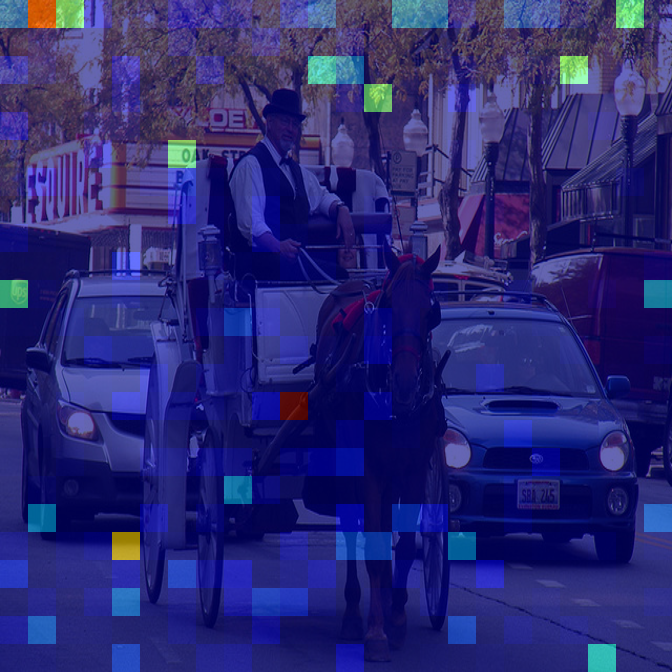}
\end{minipage} &
\begin{minipage}[t]{0.19\textwidth}\centering
\includegraphics[width=\linewidth,height=\linewidth]{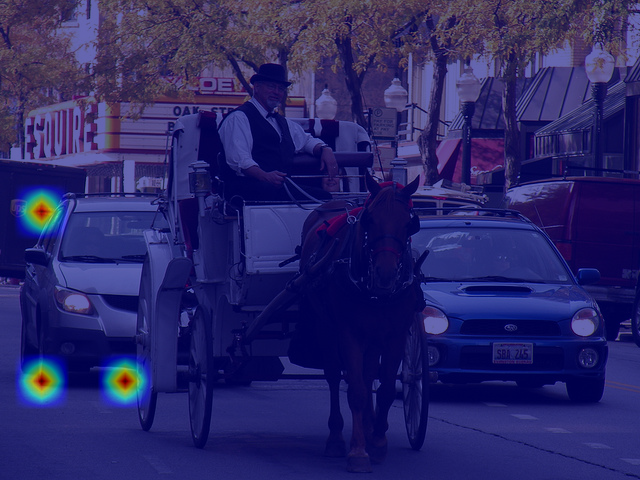}
\end{minipage} &
\begin{minipage}[t]{0.19\textwidth}\centering
\includegraphics[width=\linewidth,height=\linewidth]{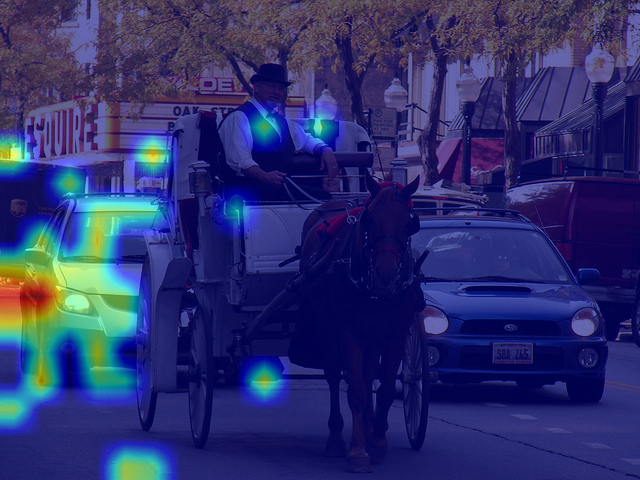}
\end{minipage} &
\begin{minipage}[t]{0.19\textwidth}\centering
\includegraphics[width=\linewidth,height=\linewidth]{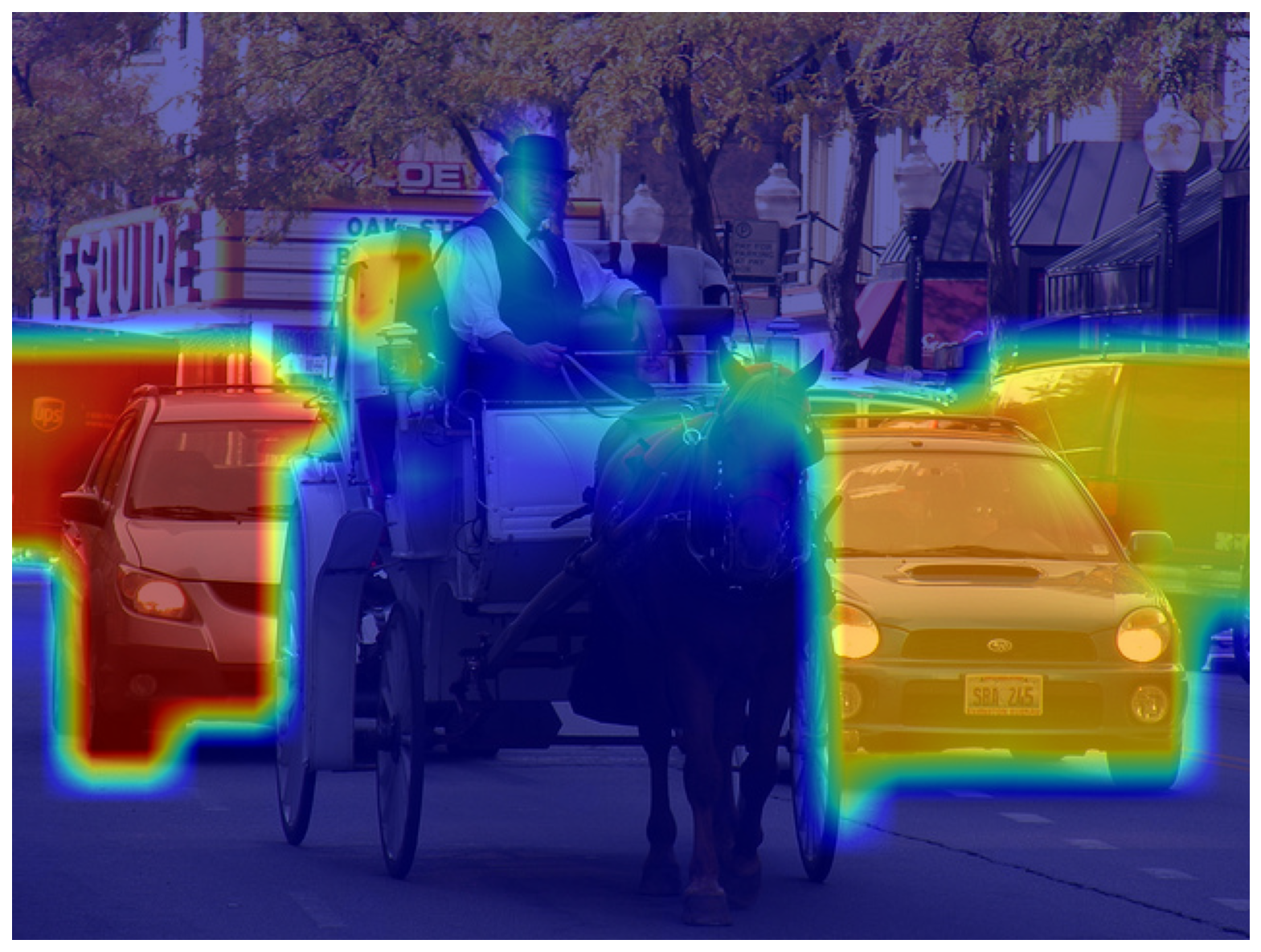}
\end{minipage} \\[2pt]

% % ================= Row 4 =================
% \begin{minipage}[t]{0.19\textwidth}\centering
% \includegraphics[width=\linewidth,height=\linewidth]{figs/4/COCO_train2014_000000391088.jpg}\par\vspace{-2pt}
% {\scriptsize Top left donut out of frame}\par\vspace{1pt}
% \end{minipage} &
% \begin{minipage}[t]{0.19\textwidth}\centering
% \includegraphics[width=\linewidth,height=\linewidth]{figs/4/raw_attn.png}
% \end{minipage} &
% \begin{minipage}[t]{0.19\textwidth}\centering
% \includegraphics[width=\linewidth,height=\linewidth]{figs/4/logitlens.png}
% \end{minipage} &
% \begin{minipage}[t]{0.19\textwidth}\centering
% \includegraphics[width=\linewidth,height=\linewidth]{figs/4/R_relevance.png}
% \end{minipage} &
% \begin{minipage}[t]{0.19\textwidth}\centering
% \includegraphics[width=\linewidth,height=\linewidth]{figs/4/heatmap_overlay_COCO_train2014_000000391088.png}
% \end{minipage} \\[2pt]

\end{tabular}

\vspace{-2mm}
\caption{
% Qualitative comparison of visual grounding heatmaps.
From left to right: (1) input image with referring expression used to query the Qwen2.5-VL model,
(2) attention map from the text-to-vision attention,
(3) Logit Lens heatmap,
(4) heatmap produced by our relevance measure $R_L$ on the base model,
and (5) heatmap produced by $R_L$ after fine-tuning with NTP+AOL.
Columns (2)–(4) are obtained from the  base model.
Our relevance measure yields more localized and semantically aligned heatmaps than attention and Logit Lens,
and fine-tuning with AOL further sharpens object localization. 
Warmer areas
indicate higher relevance.
}
\label{fig:qual_3x5_square}
\vspace{-6mm}
\end{figure*}

In comparison to Logit Lens, 
$R_L(t,v)$ differs in that it computes the normalized dot product, while
Logit Lens logits are computed as an unnormalized dot product $U^\top h_L(v)$,
which magnifies the influence of high magnitude tokens (sink tokens) that are documented to be present in VLMs \cite{barbero2025llmsattendtoken,gu2025attentionsinkemergeslanguage}.
Since the heat map values are scaled for visualization, sink tokens dominate the visualization. 
Moreover, softmax normalization
(included in the computation of $P_\theta$
in \eqref{eq:LogitLens}) magnifies the influence of the sink tokens.
Therefore, we drop the softmax normalization in the definition of $R_L$.
So, by focusing on the angle \( \angle\big(U_t, h_L(v)\big)\),
$R_L$ eliminates the influence of high norm tokens, which often heavily distorts the heat maps of Logit Lens.
As our results in Fig.~\ref{fig:qual_3x5_square} demonstrate, these differences lead to significantly better visualizations.

Our relevance measure differs fundamentally from computing cosine similarity between patch-level visual features and a sentence-level text representation in CLIP \cite{radford2021learning}. CLIP-based grounding exploits the alignment learned during contrastive pretraining, where similarity maps provide a form of localization that is external to the vision-language model and remains fixed once visual features are extracted.
In contrast, our approach defines relevance on visual tokens after they have been processed and mixed with language tokens inside the LLM.
By operating in the unembedding space of the language model, our relevance measure reflects how localized visual information is represented within the model’s internal process.

% AOL cannot be applied directly to Logit Lens, since increasing the probability $p_{\theta}(t \mid h_l(v))$ of $t$, due to the nature of softmax, may decrease the probability of generating synonyms of $t$, 

% \vspace{-3mm}

\subsection{Angular Object-Aware Loss}
% \vspace{-1mm}

Rather than training a separate decoder to predict masks~\cite{rasheed2024glamm,lai2024lisa}, we supervise
visual embeddings $h_L(v)$ themselves, in the same unembedding-aligned
space used by $R_L$. This keeps the supervision parameter-free with respect to
the architecture and avoids the forgetting risk such pathways can introduce~\cite{wu2025f}.
Our goal is to link these embeddings to the language tokens describing their content, testing whether object-level grounding can live inside the VLM's own representations.

One natural way to do this is to utilize Logit Lens.
Even though Logit Lens was initially designed for LLMs it also applies to VLMs,
as demonstrated in \cite{LogitLensICLR2025}.
It provides good-quality object confidence maps by visualizing the probability of text tokens over all visual tokens.
However, due to softmax normalization of logits,
We face two problems if we want to use it as an auxiliary loss function to improve visual grounding in the vocabulary probability space:
\begin{enumerate}
    \item if we increase the prob $p_\theta \big(t \mid h_L(v)\big)$ of a text token $t$ for a given visual token $v$, its synonyms are suppressed, since softmax is applied to logits, e.g., if $t$ represents "cat", increasing $p_{\theta}(t \mid h_l(v))$ may decrease the probability of generating "kitten".
    \item to produce a heatmap $\{ R_L(t,v) | v \in V \}$, we go over all visual tokens for a given text token $t$. But if a token $v$ is an outlier (has a high norm), its $p_\theta \big(t \mid h_L(v)\big)$ may be large, even though $t$ is irrelevant to the image content, i.e., the angle between $h_L(v)$ and $v$ is large for all $v \in V$, where $V$ is the set of the image patch tokens. 
\end{enumerate}

To address these issues, we explore a different approach.
Let text token $t$ represent the content of visual token $v$.
Our idea is to align $h_L(v)$ with $U_t$ in the embedding space.
The most obvious way to do so is to maximize their cosine similarity,
as in the definition of $R_L$ for visualization in Eq.~\eqref{eq:relevance}.

% However, 
% directly optimizing Eq.~\eqref{eq:relevance} poses a serious problem for the gradient descent, since any two vectors
% $u,x  \in \mathbb{R}^{d}$ have a very high probability of being orthogonal if $d$ is large.
% This is due to the “dead zone of cosine” that refers to the regime where optimizing objectives based on cosine similarity (or angle) becomes ineffective because the gradient provides very weak directional \cite{Deng_2022}.
% In other words,
% for $\alpha \approx \frac{\pi}{2}$,
% $\cos(\alpha) \approx 0$ 
% and barely changes when $\alpha$ changes slightly, making the loss landscape flat, causing gradient descent to get practically stuck.
% This is already for typical VLM embedding dimensions, e.g., $d=3584$
% or $d=4096$.

% To tackle this problem, we employ a margin loss that translates the cosine curve.

However, a plain cosine objective $(1-\cos\alpha)/2$, with
$\alpha=\angle(U_t,h_L(v))$, has a gradient of magnitude
$\sin\alpha/(2\|h_L(v)\|)$, which vanishes as $\alpha\to 0$: once a token
is roughly aligned, the pull toward tight alignment fades.

To address this, and following additive angular-margin
losses~\cite{Deng_2022}, we impose a stricter alignment requirement through
an angular margin. The \textbf{angular-margin similarity} is defined as
\begin{align}
ams(u,x) &=\cos(\arccos(\cos(u,x))+m)\\
&= \cos(u,x) \cos m - \sqrt{1 - \cos(u,x)^{2}}\,\sin m \nonumber
\end{align}
where $m$ is a fixed margin (typically 
$0<m<\frac{\pi}{2}$). We set $m=0.5$; its choice is analyzed in the appendix.
The range of $ams(u,x) \in [-1, cos(m)]$.
So, if we define
\begin{equation}
g(u, x) = \frac{-ams(u, x)+\cos(m)}{\cos(m)+1}, 
\end{equation}
the values of $g$ are in the interval [0, 1] with small angles between two vectors yielding values close to zero.

Let $\mathcal{P}$ denote the set of all visual (patch) tokens in the image and 
$\mathcal{P'} \subseteq \mathcal{P}$ the subset of patches overlapping with a ground-truth region
of the queried object.
In other words, we obtain $\mathcal{P'}$ by projecting the ground-truth segmentation mask onto
the visual-token grid and selecting tokens whose receptive fields overlap with
the annotated object region.

Let $ L$ be the last layer of the LLM module, $t$ be a text token, and $U_t$ be the $t$-row of the unembedding matrix $U_\theta$. We define \textbf{Angular Object-aware Loss (AOL)} as
\begin{align}
\label{eq:AOL}
\mathcal{L}_{\mathrm{AOL}}(\theta) &= 
\frac{1}{|\mathcal{P'}|} \sum_{v \in \mathcal{P'}}
g(U_t, h_L(v)) \\
&+ \frac{1}{|\mathcal{P}\setminus \mathcal{P'}|} \sum_{v \in \mathcal{P} \setminus \mathcal{P'}}
(1 - g(U_t, h_L(v))), \nonumber 
\end{align}
where $\theta$ is a vector of model parameters.
So, for each patch $v \in \mathcal{P'}$ overlapping with the queried object described by text token $t$,
we want to minimize the angle between its deep embedding and those of $t$.
At the same time, we want to increase the angle between the deep embedding of $v$ and $t$
if patch $v$ does not overlap with the queried object.
We stress that $\mathcal{L}_{\mathrm{AOL}}$ is applied only to deep embeddings of visual tokens at the final layer $L$
as it is aligned with the Unembedding matrix.
Finally, we propose to augment the standard NTP loss with:
\begin{equation}
  \mathcal{L}(\theta) = \mathcal{L}_{\mathrm{NTP}}(\theta)+ \lambda \mathcal{L}_{\mathrm{AOL}}(\theta).  
\end{equation}

\begin{table*}[hbtp]
\centering
\small
\setlength{\tabcolsep}{3pt}
\begin{tabular}{l
                @{\hspace{2pt}}>{\hspace{4pt}}p{2.9cm}<{\hspace{4pt}}@{\hspace{2pt}}
                @{\hspace{2pt}}c@{\hspace{6pt}}
                |ccc|ccc|cc}
\toprule
\textbf{Methods} & \textbf{Backbone} & \textbf{With SAM} &
\multicolumn{3}{c|}{\textbf{RefCOCO}} &
\multicolumn{3}{c|}{\textbf{RefCOCO+}} &
\multicolumn{2}{c}{\textbf{RefCOCOg}} \\
\cmidrule(lr){4-6}\cmidrule(lr){7-9}\cmidrule(lr){10-11}
& & & val & testA & testB & val & testA & testB & val & test \\
\midrule

Cropping~{\tiny\cite{yu2023zeroshotreferringimagesegmentation}}             & CLIP            & Y & 22.7 & 21.1 & 23.1 & 24.1 & 22.4 & 23.9 & 28.7 & 27.5 \\
% Global-Local CLIP~{\tiny\cite{yu2023zeroshotreferringimagesegmentation}}     & CLIP           & N     & 24.9 & 23.6 & 24.7 & 26.2 & 24.9 & 25.8 & 31.1 & 31.0 \\
TAS~{\tiny\cite{suo2023textaugmentedspatialawarezeroshot}}                   & CLIP            & N    & 29.5 & 30.3 & 28.2 & 33.2 & 38.8 & 28.0 & 35.8 & 36.2 \\
% CaR~{\tiny\cite{sun2024cliprnnsegmentcountless}}                 & CLIP            & $\times$     & 33.6 & 35.4 & 30.5 & 34.2 & 36.0 & 31.0 & 36.7 & 36.6 \\
Ref-Diff~{\tiny\cite{ni2023refdiffzeroshotreferringimage}}             & Stable Diff            & Y     & 35.2 & 37.4 & 34.5 & 35.6 & 38.7 & 31.4 & 38.6 & 37.5 \\
RESAnything~{\tiny\cite{wang2025resanythingattributepromptingarbitrary}}          & Pixtral-12B   & Y & 68.5 & 72.2 & 70.3 & 60.7 & 65.6 & 52.2 & 60.1 & 60.5 \\
Logit Lens (Qwen adapt.) ~{\tiny\cite{LogitLensICLR2025}}          & Qwen2.5-VL-7B   & N & 63.2 & 64.8  & 60.1 & 52.4 & 57.1  & 45.9 & 53.2 & 54.1 \\
Logit Lens (Qwen adapt.) + SAM2 ~{\tiny\cite{LogitLensICLR2025}}          & Qwen2.5-VL-7B   & Y & 71.3 & 72.5 & 66.8 & 60.1 & 63.5  & 53.2 & 62.8 & 64.2 \\
Localization-Head~{\tiny\cite{kang2025your}}    & LLaVA-1.5-7B   & Y & 74.2 & 76.5 & 70.4 & 62.5 & 65.2 & 56.0 & 64.2 & 68.1 \\
Localization-Head~{\tiny\cite{kang2025your}} & LLaVA-1.5-13B  & Y & \textcolor{blue}{76.1} & \textcolor{blue}{78.9} & \textcolor{blue}{72.8} & 64.1 & 67.1 & \textcolor{blue}{57.3} & \textcolor{blue}{67.7} & \textcolor{blue}{69.0} \\
\bottomrule
\multicolumn{11}{l}{\textit{Ours (training-free)}} \\
\quad Heatmap + Thr & LLaVA-1.5-7B & N & 64.6 & 66.3 & 61.2 & 54.1 & 58.8 & 47.2 & 55.0 & 55.8 \\
\quad Heatmap + SAM2                & LLaVA-1.5-7B & Y & 75.1 & 77.2 & 71.4 & \textcolor{blue}{64.5} & \textcolor{blue}{68.2} & 56.3 & 65.2 & 66.4 \\
\quad Heatmap + Thr & Qwen2.5-VL-7B & N & 66.2 & 68.4 & 62.8 & 55.3 & 60.1 & 48.6 & 56.4 & 57.2 \\
\quad Heatmap + SAM2                & Qwen2.5-VL-7B & Y & \textcolor{red}{77.4} & \textcolor{red}{79.2} & \textcolor{red}{74.8} & \textcolor{red}{67.5} & \textcolor{red}{70.5} & \textcolor{red}{59.4} & \textcolor{red}{68.6} & \textcolor{red}{69.5} \\

\bottomrule
\end{tabular}
\caption{Referring Expression Segmentation result of \textbf{training-free methods} in cIoU (\%). For our approach, \textbf{Heatmap+Thr} obtains masks by thresholding (binarizing) the relevance heatmap, while
\textbf{Heatmap+SAM2} uses the heatmap to generate masks via SAM2. ``Backbone'' denotes the main model family used for grounding, including CLIP-based methods and autoregressive VLMs.
``With SAM'' indicates whether SAM/SAM2 is used at any stage of the pipeline (Y: yes, N: no).
%to generate or refine segmentation masks. 
Red/blue denote best/second-best.}
\label{tab:res_training_free}
% \vspace{-6mm}

\end{table*}

\vspace{-4mm}
\section{Direct Gradient to Visual Tokens with AOL}
\label{sec:gradient}

\vspace{-2mm}
% AOL provides a direct gradient signal to visual tokens $h_L(s)$.
% The reason is that AOL applies directly to $h_L(s)$ via the unembedding matrix, whereas the NTP gradient must travel through text token attention paths, which greatly attenuate the gradient received by $s$.

% The key distinction between AOL and the standard NTP loss lies in the structure of their gradient paths. AOL is applied directly to the final-layer visual token representations via the unembedding matrix, yielding a one-step dependency
% in that 
% $\nabla_{h_L(s)} \mathcal{L}_{AOL}$ sends a direct signal to  $h_L(s)$ for each visual token $s$.
% Consequently, visual tokens receive a direct, attention-independent gradient signal whose magnitude admits a non-trivial lower bound under typical geometric
% conditions.
% In contrast, 
% \(
% \|\nabla_{h_L(s)} \mathcal{L}_{\mathrm{NTP}}\|=0
% \) for the final-layer $L$ visual-token outputs.
% The NTP loss is applied exclusively to text tokens, and any gradient reaching a visual token must propagate through text-to-vision attention paths. 
% So, \(
% \|\nabla_{h_l(s)} \mathcal{L}_{\mathrm{NTP}}\|>0
% \) for layers $l<L$.
% Moreover, text-to-vision attention paths
% are attenuated by multiplicative attention weights, resulting in substantially weaker and often negligible gradients for most visual tokens. Thus, AOL provides a fundamentally more direct and reliable supervision mechanism for preserving localized visual information.
% We provide formal evidence of this fact in the Appendix.

AOL and NTP differ in \emph{how} their gradients reach the visual tokens.
AOL is computed directly from the visual embeddings through the unembedding
matrix, so every visual token enters the loss explicitly and receives a gradient
in a single step. NTP, in contrast, is evaluated only at the text-output
positions, and the visual embeddings at that layer never participate in
next-token prediction; hence $\nabla_{h(v)}\mathcal{L}_{\mathrm{NTP}}=0$ for all visual tokens $v$.
NTP can influence a visual token only indirectly, by propagating back through
self-attention, where the signal is scaled by the text-to-vision attention
weights. Since these weights are small for most patches, the gradient that
actually reaches a visual token is strongly attenuated, so AOL delivers a more
direct and stronger signal than NTP.

% The direct path also gives AOL a well-behaved magnitude. Because AOL depends only on the angle between $U_t$ and $h(v)$, it is
% scale-invariant, i.e., $g(U_t,\gamma h(v))=g(U_t,h(v))$ for any $\gamma>0$,
% unlike NTP, which encourages norm growth through softmax. As derived in the
% Appendix, the gradient norm for an object token scales as
% $\sin(\alpha+m)\big/\big((\cos m+1)\,\|h(v)\|\big)$, where
% $\alpha=\angle(U_t,h(v))$. The $1/\|h(v)\|$ factor gives larger updates to
% low-norm tokens and smaller updates to high-norm ones, counteracting the
% norm-inflation pressure of NTP and matching the reduction of sink tokens in
% Fig.~\ref{fig:hidden_norms}. The margin $m$ keeps $\sin(\alpha+m)$ away from zero
% even though visual and vocabulary directions are nearly orthogonal in high
% dimensions, so the signal does not collapse into the flat ``dead zone'' of the
% cosine. In contrast, NTP admits no input-independent lower bound, since its
% visual-token gradient vanishes whenever the relevant attention weights are
% small. We provide the full derivation in the Appendix.
The direct path also gives AOL a well-behaved magnitude. Because AOL
depends only on the angle between $U_t$ and $h(v)$, it is scale-invariant,
i.e., $g(U_t,\gamma h(v))=g(U_t,h(v))$ for any $\gamma>0$. As derived in
the Appendix, the gradient norm for an object token is
$|\sin(\alpha+m)|\big/\big((\cos m+1)\,\|h(v)\|\big)$, where
$\alpha=\angle(U_t,h(v))$. The $1/\|h(v)\|$ factor gives low-norm tokens
larger angular updates than high-norm ones. Empirically, NTP+AOL also
yields fewer high-norm visual tokens (Fig.~\ref{fig:hidden_norms}); we
report this as an observation, since AOL's gradient is orthogonal to
$h(v)$ and does not directly shrink its norm. The margin keeps the gradient
nonzero near alignment, since $\sin(\alpha+m)\to\sin m>0$ as
$\alpha\to 0$, so foreground tokens continue to be pulled toward tight
alignment. In contrast, NTP admits no input-independent lower bound, since
its visual-token gradient vanishes whenever the relevant attention weights
are small. We provide the full derivation in the Appendix.

\vspace{-3mm}
\section{Experimental Results}
\vspace{-2mm}

\subsection{Tasks and Evaluation Protocol}
\vspace{-2mm}

% We evaluate on the RefCOCO, RefCOCO+, and RefCOCOg datasets~\cite{yu2016modeling}, all built on MS COCO images and requiring localization of an object specified by a natural language expression. We train on the standard training splits and report results on the validation sets.

% We first evaluate referring expression segmentation (RES) to directly assess the quality of our relevance maps and the effect of the Angular Object-aware Loss (AOL) on two VLM backbones: LLaVA-1.5-7B and Qwen2.5-VL-7B. For RES, we compute relevance heatmaps from final-layer visual token embeddings, and obtain segmentation masks either by thresholding the heatmap or by extracting a bounding box from the heatmap and using it as a prompt to a frozen segmentation model for mask refinement. No segmentation head is trained. \textcolor{red}{also REC in appendix}

We evaluate referring-expression grounding on RefCOCO/+/g~\cite{yu2016modeling}, all built on MS-COCO images and annotated with natural-language expressions for target objects. We train on the standard training splits and report results on the validation and test sets. To evaluate cross-dataset generalization, we also test the RefCOCO-trained Qwen2.5-VL model on ImageNet-S50~\cite{Gao_2023}, a category-level semantic segmentation benchmark with different images and a broader set of object categories.

For RefCOCO/+/g, we evaluate referring expression segmentation (RES) to assess the quality of our relevance maps and the effect of Angular Object-Aware Loss (AOL) on two VLM backbones: LLaVA-1.5-7B and Qwen2.5-VL-7B. The model first produces an Angular Relevance heatmap from the final-layer visual-token embeddings for the queried expression. We then convert this heatmap into a segmentation mask in two ways. In the thresholding setting, we directly binarize the heatmap to obtain a mask. In the SAM2-refinement setting, we extract a bounding box from the largest connected component of the heatmap and use that box as a prompt to a frozen SAM2 model, which refines the spatial extent of the predicted object. In both cases, semantic localization comes from our heatmap, and no segmentation head is trained. For ImageNet-S50, we use the same heatmap-to-mask procedure, except that the query is the ImageNet class name rather than a referring expression. Thresholding details are provided in the Appendix.

Segment Anything Model 2 (SAM2)~\cite{ravi2024sam2segmentimages} is applied only as a frozen post-processing module that refines spatial extents within a given box. It is not conditioned on text and does not perform object recognition, so segmentation quality reflects the semantic accuracy of the localization provided by our heatmaps. We do not use text-conditioned models such as SAM3 \cite{carion2026sam3segmentconcepts} for evaluation because they would add an external text-grounding module and are designed for simple concept prompts such as ``brown dog'', whereas RefCOCO expressions often include attributes, spatial relations, and instance-level descriptions.

\vspace{-3mm}

\subsection{Segmentation Results}
\vspace{-2mm}

In order to demonstrate the visual grounding ability of the proposed framework, 
we show the segmentation results obtained from our heatmaps.

\textbf{Training-free RES.} Training-free segmentation results for the RES task are shown in Table~\ref{tab:res_training_free}. Our relevance-based heatmaps consistently outperform maps derived from attention and Logit Lens across all dataset splits and backbones, with especially large gains over Logit Lens with Qwen2.5-VL. Interestingly, our relevance maps alone outperform CLIP-based methods. \textbf{Overall, angular relevance in the unembedding space provides a stronger grounding signal than attention or Logit Lens probabilities.} While localization-head methods, such as \cite{kang2025your}, explicitly search for or reweight specific attention heads using additional criteria, our approach simply achieves strong localization using only angular alignment between visual embeddings and vocabulary directions, without head selection or architectural modification. Moreover, strong visual grounding in our heatmaps enables effective mask refinement when combined with SAM2, yielding further improvements over direct thresholding and outperforming training-free SOTAs.

% \paragraph{RES with AOL supervision} Furthermore, results for supervised VLM-based methods are reported in Table~\ref{tab:res_finetuned}. In some cases, despite not training any segmentation heads, our thresholding-based approach achieves performance comparable to grounding models that rely on task-specific segmentation heads, highlighting the effectiveness of preserving object level information directly in visual tokens. This holds for both LLaVA-1.5-7B and Qwen2.5-VL-7B backbones, indicating that the gains from AOL are not model-specific. For fair comparison, F-LMM is evaluated using a frozen LLaVA-1.5-7B backbone, and for GLaMM we report results from the GLaMM-FS-7B (FullScope) variant. When combined with SAM2 as post-processing, our heatmaps further improve segmentation quality and reach performance comparable to state of the art, showing that strong visual grounding learned through AOL can be effectively converted into accurate masks without architectural modifications to the VLM.

\textbf{RES with AOL supervision.} Supervised VLM-based RES results are reported in Table~\ref{tab:res_finetuned}. While prior methods commonly use mask annotations to train segmentation heads, mask decoders, or additional grounding modules, we use the masks only to define positive and negative visual tokens for AOL supervision. With this representation-level supervision, NTP+AOL produces strong threshold-based masks on both LLaVA-1.5-7B and Qwen2.5-VL, without adding a trainable segmentation head. This indicates that the gains are not backbone-specific and that object-level information can be preserved directly by the VLM's visual-token representations. For fair comparison, F-LMM is evaluated using a frozen LLaVA-1.5-7B backbone, and for GLaMM we report results from the GLaMM-FS-7B FullScope variant. When heatmap-derived boxes are passed to frozen SAM2 for object-boundary refinement, performance improves further and reaches near state-of-the-art results across RefCOCO/+/g splits. These results show that \textbf{AOL learns semantically accurate localization cues that can be converted into high-quality masks} either by simple thresholding or by frozen object-boundary refinement, without architectural modification to the VLM.

\begin{table}[hbtp]
\centering
\small
\setlength{\tabcolsep}{1.0pt}
\begin{tabular}{p{3.05cm} ccc ccc cc}
\toprule
\textbf{Methods} &
\multicolumn{3}{c}{\textbf{RefCOCO}} &
\multicolumn{3}{c}{\textbf{RefCOCO+}} &
\multicolumn{2}{c}{\textbf{RefCOCOg}} \\
\cmidrule(lr){2-4}\cmidrule(lr){5-7}\cmidrule(lr){8-9}
& val & testA & testB & val & testA & testB & val & test \\
\midrule
LISA-7B~{\tiny\cite{lai2024lisa}}   & 74.1 & 76.5 & 71.1 & 62.4 & 67.4 & 56.5 & 66.4 & 68.5 \\
GSVA-7B~{\tiny\cite{xia2024gsvageneralizedsegmentationmultimodal}} & 76.4 & 77.4 & 72.8 & 64.5 & 67.7 & 58.6 & 71.1 & 72.0 \\
F-LMM-7B~{\tiny\cite{wu2025f}} 
& 75.2 & 79.1 & 71.9 & 63.7 & 71.8 & 54.7 & 67.1 & 68.1 \\
LISA-13B~{\tiny\cite{lai2024lisa}}  & 73.4 & 76.2 & 69.5 & 62.3 & 66.6 & 56.3 & 68.2 & 68.5 \\
GSVA-13B~{\tiny\cite{xia2024gsvageneralizedsegmentationmultimodal}} & 77.7 & 79.9 & 74.2 & 68.0 & 71.5 & 61.5 & 73.2 & 73.9 \\
GLaMM~{\tiny\cite{rasheed2024glamm}} & 79.5 & 83.2 & 76.9 & \textcolor{red}{75.9} & \textcolor{red}{78.7} & \textcolor{blue}{68.8} & \textcolor{red}{76.8} & \textcolor{red}{78.4} \\
PSALM~{\tiny\cite{zhang2024psalm}}     & \textcolor{red}{83.6} & \textcolor{red}{84.7} & \textcolor{red}{81.6} & 72.9 & 75.5 & \textcolor{red}{70.1} & 73.8 & 74.4 \\
\bottomrule
\multicolumn{9}{l}{\textit{Ours (NTP+AOL), backbone: LLaVA-1.5-7B}} \\
\quad Heatmap + Thr & 69.1 & 72.4 & 65.8 &  63.1 & 65.5 & 57.8 & 64.4 & 66.2 \\
\quad Heatmap + SAM2 & 78.4 & 80.6 & 74.9 & 68.5  & 71.8 & 61.2 & 69.8 & 71.2 \\

\multicolumn{9}{l}{\textit{Ours (NTP+AOL), backbone: Qwen2.5-VL-7B}} \\
\quad Heatmap + Thr  & 73.4  & 75.8 & 70.2 & 65.1 & 68.8 & 59.6 & 68.1 & 68.9 \\
\quad Heatmap + SAM2  & \textcolor{blue}{82.7}  & \textcolor{blue}{84.1} & \textcolor{blue}{79.6} & \textcolor{blue}{74.5} & 77.6  & 68.5 & \textcolor{blue}{75.4} &  \textcolor{blue}{76.5}\\

\bottomrule
\end{tabular}
% \caption{Referring Expression Segmentation results of \textbf{supervised VLM-based methods}. 
% All methods use pixel-level supervision, either by fine-tuning a VLM or by training a task-specific segmentation head.
% }
\caption{Referring Expression Segmentation results of \textbf{supervised
VLM-based methods}. All methods use pixel-level supervision, either by
fine-tuning a VLM or by training a task-specific segmentation head.
\textbf{AOL finetuning improves average RES cIoU over the base model by
+7.7 (LLaVA, Thr), +4.0 (LLaVA, SAM2), +9.4 (Qwen, Thr), and +6.5 (Qwen,
SAM2) points.}}
\label{tab:res_finetuned}
% \vspace{-4mm}
\end{table}

As shown in Fig.~\ref{fig:qual_3x5_square}, our relevance-based ($R_L$) heatmaps provide substantially clearer and more accurate object localization than both attention maps and Logit Lens, which often highlight imprecise or semantically mixed regions. In contrast, $R_L$ produces compact responses aligned with the referred objects. Adding AOL on top of NTP further sharpens localization and improves instance discrimination. For example, in the third row, the model activates on multiple objects of the same category (cars) but assigns the highest relevance to the specific instance described by the referring expression (the left gray car).

% Moreover, when bounding boxes derived from our heatmaps are used to prompt SAM2 (Fig.~\ref{fig:segment}), the resulting masks closely match the ground-truth segmentation, further confirming that AOL improves the quality of visual grounding encoded in visual tokens.

% \paragraph{Cross-dataset generalization on ImageNet-S50.}
% We evaluate how Qwen2.5-VL trained with NTP+AOL on RefCOCO, whose images and object annotations come from MS-COCO, performs on another dataset. Specifically, we evaluate whether the learned visual representations transfer to ImageNet-S50 \cite{Gao_2023}, which contains different images and a broader set of object categories. Since some ImageNet-S50 classes still have semantic overlap with MS-COCO, we additionally report results on the 24 COCO-disjoint classes, defined as classes with no corresponding MS-COCO category. The class partitioning procedure and full mapping are provided in Appendix~\ref{app:imagenet_s50_partition}. miou - > thresholded mask, EPG - > raw heatmaaps\textcolor{red}{ TO be continued ... }

\begin{figure}[hbtp]
    \centering
    \includegraphics[width=0.9\columnwidth]{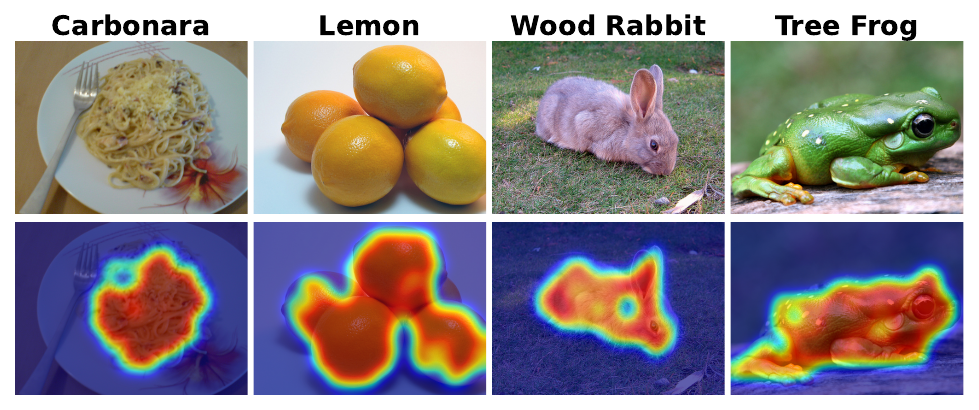}
    \caption{Angular Relevance heatmaps on COCO-disjoint ImageNet-S50 classes. The model localizes object categories absent from RefCOCO training, including food items and animals.}
    \label{fig:imagenet_s50_ood}
    % \vspace{-3mm}
\end{figure}

\textbf{Cross-dataset generalization on ImageNet-S50.}
We evaluate whether the visual representations learned by Qwen2.5-VL with NTP+AOL on RefCOCO transfer to ImageNet-S50~\cite{Gao_2023}, which contains different images and a broader set of object categories. To determine the generalization ability of our approach to unfamiliar objects, we partition ImageNet-S50 into COCO-overlap classes, which have a similar MS-COCO category, and COCO-disjoint classes, which have no corresponding MS-COCO category. The full mapping is provided in Appendix section “ImageNet-S50 Class Partition”. 
% Table~\ref{tab:imagenet_s50_ood} reports mIoU from thresholded heatmaps. Even before AOL training, Angular Relevance outperforms Logit Lens by 8.1 mIoU on all classes and 6.0 mIoU on COCO-disjoint classes.
Table~\ref{tab:imagenet_s50_ood} reports mIoU from thresholded heatmaps. Even before AOL training, Angular Relevance outperforms Logit Lens, and NTP-only yields only minor gains; in contrast, NTP+AOL improves Angular Relevance from 36.9 to 57.4 mIoU on all classes and from 33.5 to 51.6 mIoU on COCO-disjoint classes. 
% After NTP+AOL training, Angular Relevance further improves from 36.9 to 57.4 mIoU on all classes and from 33.5 to 51.6 mIoU on COCO-disjoint classes.
\textbf{The disjoint-class gain (+18.1 mIoU) is only modestly smaller than the overlap-class gain (+22.6 mIoU), suggesting the generalization ability of AOL to unfamiliar out-of-distribution objects.} Fig.~\ref{fig:imagenet_s50_ood}
shows qualitative examples on COCO-disjoint classes.

%shows that AOL improves visual-token localization beyond the MS-COCO category vocabulary used during RefCOCO training.

% \begin{table}[hbtp]
% \centering
% \caption{Localization results on ImageNet-S50, evaluated on all
% 50 classes and on the 24 classes with no corresponding MS-COCO
% category (COCO-disjoint).}
% \label{tab:imagenet_s50_ood}
% \small
% \setlength{\tabcolsep}{3pt}
% \renewcommand{\arraystretch}{1.08}
% \begin{tabular}{@{}llcccc@{}}
% \toprule
% Method & Setting
% & \multicolumn{2}{c}{All 50}
% & \multicolumn{2}{c}{CD 24} \\
% \cmidrule(lr){3-4}\cmidrule(lr){5-6}
% & & mIoU & EPG  & mIoU & EPG \\
% \midrule
% Logit Lens
% & Base model    & 28.8 & 34.2 & 27.5 & 32.1 \\
% % & NTP+AOL & 41.3 & 66.3 & 39.8 & 61.7 \\
% \midrule
% Ang. Rel.
% & Base model   & 36.9 & 39.0 & 33.5 & 38.2 \\
% & NTP+AOL & \textbf{57.4} & \textbf{74.5} & \textbf{51.6} & \textbf{67.6} \\
% \bottomrule
% \end{tabular}
% \end{table}

\begin{table}[hbtp]
\centering
\small
\setlength{\tabcolsep}{3pt}
\renewcommand{\arraystretch}{1.08}
\begin{tabular}{@{}llccc@{}}
\toprule
Method & Setting & All & Overlap & Disjoint \\
\midrule
Logit Lens & Base & 28.8 & 29.9 & 27.5 \\
& NTP & 30.2 & 31.2 & 28.7 \\

\midrule
Ang. Rel. & Base & 36.9 & 39.8 & 33.5 \\
& NTP & 38.5 & 41.6 & 35.2 \\

& NTP+AOL & \textbf{57.4} & \textbf{62.4} & \textbf{51.6} \\
\cmidrule(lr){1-5}
\multicolumn{2}{l}{\textit{$\Delta=$ (NTP+AOL $-$ Base)}} & \textit{+20.5} & \textit{+22.6} & \textit{+18.1} \\
\bottomrule
\end{tabular}
\caption{ImageNet-S50 cross-dataset generalization. Results in mIoU for Qwen2.5-VL trained on RefCOCO and evaluated on ImageNet-S50. Overlap classes have some similarity to MS-COCO classes, whereas Disjoint classes do not.}
\label{tab:imagenet_s50_ood}
% \vspace{-7mm}
\end{table}

More experimental results, ablations, visualizations, and implementation details are provided in the appendix.

\vspace{4mm}
\section{Ablation Studies}
% \vspace{-2mm}

\textbf{Causal faithfulness.}
To test whether AR identifies patches that the model actually uses, we perform a representation-intervention experiment on RefCOCO val. We first set $k$ to the number of visual patches corresponding to the ground-truth object region for each image. Replacing the top-$k$ AR patches with the average patch embedding reduces referring expression comprehension accuracy from \textbf{91.8\% to 77.4\%}, whereas replacing the same number of random patches reduces it only to \textbf{87.6\%}. Thus, patches assigned high relevance by AR are substantially more important to the model's prediction than random patches.

\begin{figure}[hbtp]
\centering
\includegraphics[width=\columnwidth]{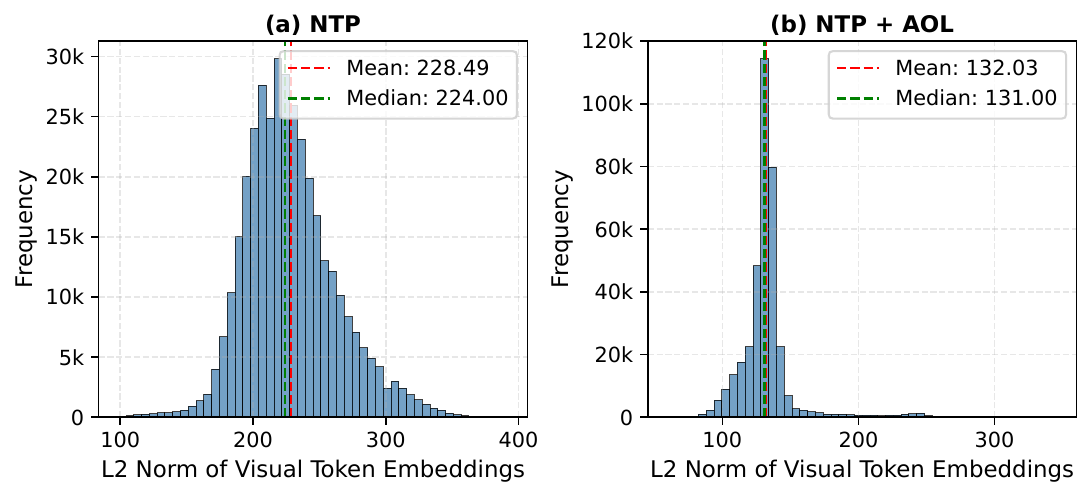}
\caption{$L_2$ norms of final-layer visual token embeddings on RefCOCO validation. Compared with NTP-only finetuning, NTP+AOL produces a more compact visual-token norm distribution and fewer high-norm outliers.}
\label{fig:hidden_norms}
% \vspace{-5mm}
\end{figure}

\textbf{Effect of AOL on visual embedding norm.} As demonstrated in histograms in Fig.~\ref{fig:hidden_norms},
AOL+NTP compresses the norm distribution of visual embeddings and reduces high-norm outliers, which are known to cause sink-token behavior and magnitude-dominated explanations.
This makes heatmaps derived from our relevance measure (Eq.~\eqref{eq:relevance}) more faithful and spatially localized.

% \textbf{Generalization across related text queries.}
% We also test whether AOL overfits to the exact object word used during supervision. Fig.~\ref{fig:cat_synonyms} shows Angular Relevance maps for several cat-related queries. Although AOL is trained using only the category name ``cat'', the resulting heatmaps also improve for related words such as ``kitten'', ``feline'', and ``tabby''. This suggests that AOL improves \textbf{local concept alignment rather than only the supervised word, allowing localization gains to transfer to synonymous and closely related queries}.

% \begin{figure}[hbtp]
%     \centering
%     \includegraphics[width=0.9\columnwidth]{figs/cat_synonyms.pdf}
%     \caption{Angular Relevance maps for synonym queries. Top: base model; bottom: NTP+AOL. Although AOL supervision only uses the category name ``cat'', localization also improves for related queries such as ``kitten'', ``feline'', and ``tabby''.}
%     \label{fig:cat_synonyms}
%     \vspace{-4mm}
% \end{figure}

% \textbf{General VQA performance.}
% To check whether AOL affects the general language and visual-question-answering ability of the VLM, we evaluate Qwen2.5-VL on VQAv2 \cite{goyal2017making} and POPE \cite{POPE2023}. As shown in Table~\ref{tab:vqa_performance}, NTP+AOL does not degrade performance relative to the base model or NTP-only finetuning. We observe small improvements on both VQAv2 and POPE, suggesting that the added visual token supervision with AOL is compatible with the model's general behavior.

\textbf{General VQA and hallucination.}
To assess whether AOL affects general VLM behavior, we evaluate Qwen2.5-VL on VQAv2~\cite{goyal2017making} and POPE~\cite{POPE2023}.  As shown in Table~\ref{tab:vqa_performance}, \textbf{NTP+AOL does not degrade general question answering or object hallucination} relative to the base model or NTP-only finetuning, and slightly improves both metrics.

\begin{table}[htbp]
  \centering
  \small
  \begin{tabular}{lcc}
    \toprule
    \textbf{Method} & \textbf{VQAv2} & \textbf{POPE} \\
    \midrule
    Base Model & 80.5 & 86.77 \\
    NTP        & 80.6 & 87.50 \\
    NTP + AOL  & \textbf{81.2} & \textbf{89.87} \\
    \bottomrule
  \end{tabular}
\caption{General VQA and hallucination performance for Qwen2.5-VL-7B.}
\label{tab:vqa_performance}
% \vspace{-5mm}
\end{table}

\textbf{AOL with pseudo-mask supervision.}
We also find that AOL can be trained without ground-truth masks. Starting from Qwen2.5-VL, which obtains 66.2 cIoU before AOL training, pseudo-masks generated by a specialized segmentation model can be used to define positive and negative visual tokens during training, improving RefCOCO validation cIoU to 69.6 and approaching the 73.4 cIoU obtained with ground-truth masks. Details are provided in the appendix section ``AOL with Pseudo-mask Supervision.''

% \vspace{-4mm}
\section{Related Work}
Large-scale VLMs align images and text via contrastive learning, frozen vision encoders, or autoregressive LMs~\cite{radford2021learning,alayrac2022flamingo}. We focus on instruction-tuned autoregressive VLMs~\cite{li2023blip,liu2023visual,bai2023qwenvlversatilevisionlanguagemodel}, which perform well on captioning, retrieval, and VQA~\cite{li2023blip,wang2021simvlm,radford2021learning,yu2022coca,alayrac2022flamingo}. However, they can hallucinate, underuse visual tokens, and exhibit attention-sink behavior~\cite{esmaeilkhani2026logitlenssupervisionpatchlevel,payattentionimageECCV2024,chen2024imageworth12tokens,kang2025attentionsink, ansari2026erunderstandevaluatingvisionlanguagemodels}, suggesting that NTP weakly supervises patches and allows visual embeddings to drift from image content. Also, removing visual tokens inside ground-truth regions strongly reduces performance~\cite{neo2025interpretingvisualinformationprocessing}.

% Several lines of work address grounding at different levels. Segmentation-oriented VLMs such as LISA~\cite{lai2024lisa}, F-LMM~\cite{wu2025f}, and GLAMM~\cite{rasheed2024glamm} introduce grounding-specific mechanisms for mask or region prediction, while referring-expression methods often use task-specific architectures for box or mask grounding~\cite{mao2016generation,yu2016modelingcontextreferringexpressions}. Other approaches formulate localization as coordinate or token prediction, as in Pix2Seq and Kosmos-2~\cite{pix2seqICLR2022,kosmos2-ICLR2023}, or learn image-text cross-modal alignment, as in CLIP, UNITER, and ViLBERT~\cite{radford2021learning,chen2020uniter,lu2019vilbert}. In contrast, AOL targets grounding inside the VLM by aligning internal visual-token embeddings with object-specific vocabulary directions.

% Recent works regularize VLMs to preserve visual information during instruction tuning. VIRAL aligns internal visual representations with pretrained vision-foundation-model features, while Reconstructive Visual Instruction Tuning uses reconstruction-based supervision~\cite{yoon2025visualrepresentationalignmentmultimodal,wang2024reconstructivevisualinstructiontuning}. AOL instead applies object-level angular supervision in the LLM unembedding space, aligning visual tokens with vocabulary directions after cross-modal mixing.
Several lines of work address grounding or visual preservation at different levels. Segmentation-oriented VLMs such as LISA~\cite{lai2024lisa}, F-LMM~\cite{wu2025f}, and GLAMM~\cite{rasheed2024glamm} add grounding-specific mechanisms for mask or region prediction, while referring-expression methods often use task-specific architectures for box or mask grounding~\cite{mao2016generation,yu2016modeling}. Other approaches use coordinate or token prediction, as in Pix2Seq and Kosmos-2~\cite{pix2seqICLR2022,kosmos2-ICLR2023}, global cross-modal alignment, as in CLIP, UNITER, and ViLBERT~\cite{radford2021learning,chen2020uniter,lu2019vilbert}, or visual-preservation losses such as VIRAL's feature alignment with vision foundation model representations and ROSS's reconstruction-based supervision during visual instruction tuning~\cite{yoon2025visualrepresentationalignmentmultimodal,wang2024reconstructivevisualinstructiontuning}. In contrast, AOL targets grounding inside the VLM by aligning internal visual-token embeddings with object-specific vocabulary directions.

Finally, visualization methods such as attention analysis~\cite{darcet2023vision,kang2025attentionsink,esmaeilkhani2025directvisualgroundingdirecting}, VL-InterpreT~\cite{aflalo2022vlinterpretinteractivevisualizationtool}, Grad-CAM~\cite{Selvaraju_2019}, and Logit Lens~\cite{logitlens2020,LogitLensICLR2025} provide spatial or token-level cues. Angular Relevance instead evaluates how well visual embeddings align with vocabulary directions, generating heatmaps that support both analysis and grounding.

\vspace{-3mm}
\section{Conclusion}
 \vspace{-2mm}

% We proposed Angular Relevance ($R_L$) for explainability in autoregressive VLMs, producing more faithful and localized heatmaps than attention maps and Logit Lens. We further introduced Angular Object-Aware Loss (AOL), a direct angular supervision on visual embeddings that preserves localized visual information during cross-modal mixing. Experiments on RES task show strong training-free grounding and consistent gains after NTP+AOL finetuning, improving RES without training any segmentation heads. Combined with frozen SAM2 for mask refinement, our heatmaps yield near state-of-the-art segmentations, and the learned grounding transfers to object categories absent from training all while preserving general VQA performance.
We proposed Angular Relevance ($R_L$) for explainability in autoregressive VLMs, producing more faithful and localized heatmaps than attention maps and Logit Lens. We further introduced Angular Object-Aware Loss (AOL), a direct angular supervision on visual embeddings that preserves localized visual information during cross-modal mixing. Experiments on the RES task show strong training-free grounding and consistent gains after NTP+AOL finetuning, improving segmentation without training any segmentation heads. The learned grounding further transfers to object categories absent from training, all while preserving general VQA performance.

\vspace{-3mm}
\section{Acknowledgment}
\vspace{-2mm}
This work was supported by NSF grant IIS-2331768.

% \newpage

%\appendix

% \section*{Ethical Statement}

% There are no ethical issues.

% %\section*{Acknowledgments}

{
    \small
    \bibliographystyle{ieeenat_fullname}
    \bibliography{main}
}

\end{document}

\maketitle
\appendix

% AOL provides a direct gradient signal to visual tokens $h_L(s)$.
% The reason is that AOL applies directly to $h_L(s)$ via the unembedding matrix, whereas the NTP gradient must travel through text token attention paths, which greatly attenuate the gradient received by $s$.
\section{Implementation and Evaluation Details}

\paragraph{Training Prompt.}
We use the same prompt template for training and evaluation, instructing the VLM to output a single bounding box for the referred object:
\begin{quote}\small
\texttt{Given the image and the referring expression, output the bounding box coordinates (x1, y1, x2, y2) of the referred object in pixel coordinates. Output only the four numbers.}
\end{quote}

\paragraph{Heatmap-to-Mask Generation and Thresholding.} For threshold-based mask generation, we first apply ReLU to the heatmap and then binarize it using a scalar threshold computed from per-image heatmap statistics. We evaluate thresholds $\tau \in \{\mu-\sigma,\ \mu,\ \mu+\sigma\}$, where $\mu$ and $\sigma$ are the mean and standard deviation of the heatmap values for that image. Pixels with values above $\tau$ are retained in the binary mask. For Angular Relevance heatmaps, $\tau=\mu+\sigma$ gives the best segmentation performance, while for Logit Lens heatmaps, $\tau=\mu$ performs best. 
For evaluation fairness, thresholds were selected once on the RefCOCO train split (AR: $\mu+\sigma$; Logit Lens: $\mu$) and then fixed across all evaluation datasets.

\paragraph{AOL Hyperparameters.}
We ablate the AOL weight $\lambda \in \{0.25,0.5,0.75,1,1.5,2\}$ and the angular margin $m \in \{0.1,0.3,0.5,0.7\}$ radians. On a held out set from RefCOCO train with Qwen2.5-VL, $\lambda=2$ and $m=0.5$ achieve the best performance and are used in the main experiments.

\paragraph{Choice of angular margin $m$.}
The gradient magnitude of AOL scales as
$|\sin(\alpha+m)|/((\cos m+1)\|h_L(s)\|)$ (Eq.~9). As $m\to 0$, AOL
reduces to a plain cosine objective whose gradient vanishes as
$\alpha\to 0$. A larger $m$ keeps the gradient magnitude nonzero near
alignment but reduces it for approximately orthogonal tokens (at
$\alpha=\pi/2$ it equals $\cos m/((1+\cos m)\|h_L(s)\|)$, decreasing in
$m$). We swept $m\in\{0.1,0.3,0.5,0.7\}$ on RefCOCO train with
Qwen2.5-VL; $m=0.5$ performed best and is used for all datasets and both
backbones.

\paragraph{Implementation Configuration.}
We finetuned the LLava-v1.5-7B and Qwen2.5-VL-7B-Instruct VLMs using DeepSpeed on four NVIDIA A100 GPUs (80 GB VRAM each). The vision encoder remained frozen except for a trainable projection layer, while Low-Rank Adaptation (LoRA) was applied to the layers of the language model: both LLava-v1.5 and Qwen2.5-VL employed a LoRA rank of 16. Note that for the AOL, the language head, which represents the unembedding matrix, was frozen to stay consistent. Training was conducted with a per-GPU batch size of 16 and gradient accumulation of 4 (effective global batch size = 256). Optimization followed a cosine decay learning rate schedule (initial LR = $5\times10^{-5}$) with a linear warm-up over the first 3\% of total training steps. 

\section{Layer Ablation for AOL Supervision}

Since $\mathcal{L}_{\mathrm{AOL}}$ aligns visual-token embeddings with rows of
the unembedding matrix $U$, we expect supervision at deeper layers to be more
effective, because the final hidden states are directly mapped by $U$ to
vocabulary logits. To verify this, we apply AOL to different layer ranges of
Qwen2.5-VL and evaluate cIoU on RefCOCO validation using thresholded Angular
Relevance heatmaps.

\begin{table}[hbtp]
\centering
\small
\begin{tabular}{@{}lccc@{}}
\toprule
Layers & First 2 & Middle 2 & Last 1 \\
\midrule
cIoU & 67.5 & 70.3 & \textbf{73.4} \\
\bottomrule
\end{tabular}
\caption{Effect of the AOL supervision layer range on RefCOCO validation cIoU
(Qwen2.5-VL, NTP+AOL, thresholded heatmaps).}
\label{tab:layer_ablation}
\end{table}

Supervising the final layer performs best, consistent with the fact that the
unembedding matrix is most directly aligned with final-layer representations.

% \section{Effect of NTP and AOL Finetuning}

% We isolate the effect of standard next-token prediction (NTP) finetuning from
% the proposed AOL objective on Qwen2.5-VL. Table~\ref{tab:ntp_aol_ablation}
% shows that NTP alone provides only a small improvement over the base model,
% whereas adding AOL yields a substantially larger improvement in localization
% quality.

% \begin{table}[hbtp]
% \centering
% \small
% \begin{tabular}{@{}lc@{}}
% \toprule
% Model & RefCOCO val cIoU \\
% \midrule
% Base Qwen2.5-VL & 66.2 \\
% NTP & 66.8 \\
% NTP+AOL & \textbf{73.4} \\
% \bottomrule
% \end{tabular}
% \caption{Effect of NTP-only finetuning and NTP+AOL on RefCOCO validation cIoU
% (Qwen2.5-VL, thresholded Angular Relevance heatmaps).}
% \label{tab:ntp_aol_ablation}
% \end{table}

\section{AOL Component and Margin Ablations}

We isolate the effects of NTP, the angular margin, and the foreground and
background components of AOL on Qwen2.5-VL.

\begin{table}[hbtp]
\centering
\small
\begin{tabular}{@{}lc@{}}
\toprule
Model setting & RefCOCO val cIoU \\
\midrule
Base Qwen2.5-VL & 66.2 \\
NTP & 66.8 \\
{NTP+AOL ($m=0$)} & 67.9 \\
{NTP+AOL (background only)} & 68.7 \\
{NTP+AOL (foreground only)} & 71.2 \\
NTP+AOL (full) & \textbf{73.4} \\
\bottomrule
\end{tabular}
\caption{Ablation of the AOL objective on RefCOCO validation
(Qwen2.5-VL, thresholded Angular Relevance heatmaps).}
\label{tab:ntp_aol_ablation}
\end{table}

Without the angular margin ($m=0$), AOL reaches only 67.9 cIoU, showing that the margin is important in practice. 
% consistent
% with the near-orthogonal ``dead zone of cosine'' discussed in the main paper.

Foreground-only and background-only supervision reach 71.2 and 68.7 cIoU,
respectively, while combining both terms performs best at 73.4, showing that
both components contribute, with foreground alignment providing the larger
individual gain.

\section{Generalization across related text queries.}
\begin{figure}[hbtp]
    \centering
    \includegraphics[width=0.9\columnwidth]{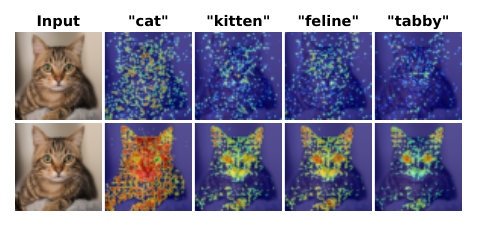}
    \caption{Angular Relevance maps for synonym queries. Top: base model; bottom: NTP+AOL. Although AOL supervision only uses the category name ``cat'', localization also improves for related queries such as ``kitten'', ``feline'', and ``tabby''.}
    \label{fig:cat_synonyms}
    \vspace{-4mm}
\end{figure}

We also test whether AOL overfits to the exact object word used during supervision. Fig.~\ref{fig:cat_synonyms} shows Angular Relevance maps for several cat-related queries. Although AOL is trained using only the category name ``cat'', the resulting heatmaps also improve for related words such as ``kitten'', ``feline'', and ``tabby''. This suggests that AOL improves \textbf{local concept alignment rather than only the supervised word, allowing localization gains to transfer to synonymous and closely related queries}.

\section{Direct Gradient to Visual Tokens with AOL}

\subsection{Lower bound of AOL gradient}

% \paragraph{We show that the norm of the AOL gradient is bounded from below by a positive constant.}
\paragraph{
We show that the AOL gradient with respect to visual tokens is strictly positive and scales as $1/\|h_L(s)\|$.}
Under bounded hidden-state norms and typical near-orthogonality between visual-token directions and vocabulary directions in high-dimensional spaces, this implies a positive lower bound for the majority of visual tokens.

For simplicity, we focus on a single positive (object), visual token $P' = \{s\}$ in the last layer $ L$. Then $\mathcal{L}_{AOL}$ reduces to 
$\mathcal{L}_{AOL} = g(U_t, h_L(s))$.
The gradient with respect to the token representation $h_L(s)$ is
\begin{align} \label{eq:grad}
&\nabla_{h_L(s)}\mathcal{L}_{AOL} = \nabla_{h_L(s)} g\big(U_t, h_L(s)\big)\\
&=
-\frac{1}{\cos m + 1}
\left(
\cos m
+
\frac{c\,\sin m}{\sqrt{1 - c^2}}
\right)
\frac{\hat U_t - c\,\widehat{h_L(s)}}{\|h_L(s)\|},
\end{align}
where 
\[
c
=
\frac{U_t^\top h_L(s)}{\|U_t\|\,\|h_L(s)\|},
\quad
\hat U_t
=
\frac{U_t}{\|U_t\|},
\quad
\widehat{h_L(s)}
=
\frac{h_L(s)}{\|h_L(s)\|}.
\]

%%%%%%%%%%% lower bound
For \(\alpha = \angle\big(U_t, h_L(s)\big)\), we can express the gradient magnitude as
% \begin{equation} \label{eq:sin}
% \left\|\nabla_{h_L(s)} g\big(U_t, h_L(s)\big)\right\|
% =
% \frac{\sin(\alpha + m)}{(\cos m + 1)\,\|h_L(s)\|}.   
% \end{equation}
\begin{equation} \label{eq:sin}
\left\|\nabla_{h_L(s)} g\big(U_t, h_L(s)\big)\right\|
=
\frac{\left|\sin(\alpha + m)\right|}{(\cos m + 1)\,\|h_L(s)\|}.   
\end{equation}

% For random vectors in high-dimensional spaces, we have
% \(
% \alpha \approx \mathcal{N}\!\left(\frac{\pi}{2}, \frac{1}{d}\right).
% \)
As a high-dimensional heuristic, angles between weakly correlated vectors concentrate near $\tfrac{\pi}{2}$ with standard deviation $O(1/\sqrt d)$.
Accordingly, we treat
\[
\alpha \approx \mathcal{N}\!\left(\tfrac{\pi}{2}, \tfrac{1}{d}\right)
\]
as an approximation to characterize typical behavior, rather than as a worst-case guarantee.

Consequently, the angular deviation from orthogonality concentrates around \(\pi/2\) with
\(
\operatorname{std} \approx \frac{1}{\sqrt{d}} \;\text{radians}
\),
e.g., for Qwen2.5-VL, \(d = 3584\), $\frac{1}{\sqrt{d}}
\approx 0.0167 \;\text{rad} < 1^\circ$.

We obtain that the majority of deep embeddings of visual tokens is orthogonal to the rows of unembedding matrix $U$, i.e.,
\(
\alpha \in \left[\frac{\pi}{2}-\varepsilon,\ \frac{\pi}{2}+\varepsilon\right]
\)
for a small $\varepsilon = \frac{1}{\sqrt{d}}$.
Over this interval, a lower bound is obtained by taking the minimum of \(\sin(\alpha + m)\).
Since \(\sin(\tfrac{\pi}{2}+z)=\cos z\),
$\min \sin(\alpha + m) = \min \{\cos(m-\varepsilon),\ \cos(m+\varepsilon)\} = \cos(m+\varepsilon)$.
So, we obtain
\begin{align} \label{eq:minest}
\frac{ \cos(m-\varepsilon)}
{(\cos m + 1)\,\|h_L(s)\|}
\ge
\left\|\nabla_{h_L(s)} g\big(U_t, h_L(s)\big) \right\| \nonumber \\
\ge
\frac{ \cos(m+\varepsilon)}
{(\cos m + 1)\,\|h_L(s)\|}.  
\end{align}
% Both the lower and upper bounds
% are positive if
% $m-\varepsilon, m+\varepsilon,  m+1 \in [0, \pi)$,
% which is easily satisfied for any autoregressive VLM at test time,
% i.e., it is independent of the input data, since $m$ is always set $m=0.5$.
% So, large-norm tokens $s$ receive smaller AOL gradients, while small-norm tokens receive stronger ones.
% This implies that AOL should reduce the number of high norm (sink tokens),
% which is demonstrated experimentally in Fig.~\ref{fig:two_by_two_singlecol}.
Since $m\in(0,\tfrac{\pi}{2})$ and $\alpha \approx \tfrac{\pi}{2}$, we have
$\alpha + m \in (0,\pi)$, and hence $\sin(\alpha+m) > 0$, so the bounds in
\eqref{eq:minest} are positive.
Thus, for a fixed angle, visual tokens with large norms receive smaller AOL
gradients than tokens with small norms. However, since AOL is scale
invariant, its gradient is orthogonal to the token embedding and does not
directly reduce its norm. Empirically, we observe that NTP+AOL reduces
outlier visual tokens with high norms, as shown in Fig.~4 of the main paper.

We have estimated that 
$\lVert h_L(s)\rVert \leq 150$ for around 95\% of visual tokens in Qwen2.5-VL.
% So, $\|\nabla_{h_L(s)}\mathcal{L}_{AOL}\|$
% is bounded below for most visual tokens $s$ when its parameters are set to practical values.
So, $\|\nabla_{h_L(s)}\mathcal{L}_{AOL}\|$
admits a positive lower bound for the majority of visual tokens $s$
under the empirically observed bound on $\|h_L(s)\|$.
Since the gradient to the lower layers $l<L$ propagates through a feed-forward network,
$\|\nabla_{h_l(s)}\mathcal{L}_{AOL}\|$ will retain strong signal for at least higher layers $l<L$.

\subsection{Absence of a non-trivial lower bound for NTP}

The norm of the next-token prediction (NTP) loss
\(
\|\nabla_{h_L(s)} \mathcal{L}_{\mathrm{NTP}}\|=0
\) for the final-layer $L$ visual-token outputs.
Hence NTP loss signal can only propagate to lower layers through self-attention with text tokens. Here we focus only on layer $L-1$.
Unlike the Object Awareness Loss (AOL), the next-token prediction (NTP) loss does not admit a non-trivial uniform lower bound on the gradient norm
\(
\|\nabla_{h_{L-1}(s)} \mathcal{L}_{\mathrm{NTP}}\|
\)
for most visual tokens \(s\).
The reason is structural: the NTP loss is applied only to text tokens, and gradients can reach a visual token \(s\) solely through attention-mediated paths.
If the attention from text tokens to \(s\) is small, then the corresponding Jacobians
\(
\partial h_{L-1}(j) / \partial h_{L-1}(s)
\)
are small for all \(j\in\mathcal{T}\), causing the NTP gradient at \(h_{L-1}(s)\) to become arbitrarily close to zero.
Therefore, without additional assumptions on attention weights, the best unconditional lower bound on
\(
\|\nabla_{h_{L-1}(s)} \mathcal{L}_{\mathrm{NTP}}\|
\)
is zero.

\section{ImageNet-S50 Class Partition}

To analyze model behavior across varying degrees of semantic overlap with
MS-COCO, we partition the 50 ImageNet-S50 classes into COCO-overlap and
COCO-disjoint groups by comparing each class name to the 80 MS-COCO categories.
The COCO-overlap group is further divided into fine-grained and loose
correspondences. Ambiguous cases were placed in the loose COCO-overlap group.
% The mapping was additionally checked with two LLM assistants
% (GPT-5.5~Thinking and Claude~Sonnet~4.6) and manually verified by the authors.
The mapping was first constructed manually, then cross-checked using GPT-5.5 Thinking and Claude Sonnet 4.6, and finally manually verified by the authors.

\paragraph{COCO-overlap: fine-grained correspondences (18/50, 36\%).}
Classes that are fine-grained instances of a COCO category
(Table~\ref{tab:exact}).

\begin{table}[hbtp]
\centering
\small
\begin{tabular}{ll}
\toprule
\textbf{ImageNet-S50 Class} & \textbf{COCO Category} \\
\midrule
Goldfinch           & bird         \\
Kuvasz              & dog          \\
Siamese cat         & cat          \\
American black bear & bear         \\
African elephant    & elephant     \\
Airliner            & airplane     \\
Beach wagon         & car          \\
Bullet train        & train        \\
Cellular telephone  & cell phone   \\
Dining table        & dining table \\
Golf ball           & sports ball  \\
Mixing bowl         & bowl         \\
Park bench          & bench        \\
Purse               & handbag      \\
Television          & tv           \\
Umbrella            & umbrella     \\
Vase                & vase         \\
Water bottle        & bottle       \\
\bottomrule
\end{tabular}
\caption{ImageNet-S50 classes with fine-grained COCO-overlap correspondences.}
\label{tab:exact}
\end{table}

\paragraph{COCO-overlap: loose correspondences (8/50, 16\%).}
Classes with a related but imprecise COCO label (Table~\ref{tab:loose}).
Some are visually in-distribution (\textit{e.g.}, motor scooter~$\to$~motorcycle),
while others share a COCO label by name but differ substantially in visual
appearance from typical COCO instances of that category
(\textit{e.g.}, container ship~$\to$~boat, digital watch~$\to$~clock).

\begin{table}[hbtp]
\centering
\small
\begin{tabular}{ll}
\toprule
\textbf{ImageNet-S50 Class} & \textbf{COCO Category} \\
\midrule
Giant panda    & bear       \\
Container ship & boat       \\
Digital watch  & clock      \\
Motor scooter  & motorcycle \\
Streetcar      & train      \\
Toilet seat    & toilet     \\
Yawl           & boat       \\
Street sign    & stop sign  \\
\bottomrule
\end{tabular}
\caption{ImageNet-S50 classes with loose COCO-overlap correspondences.}
\label{tab:loose}
\end{table}

\paragraph{COCO-disjoint classes (24/50, 48\%).}
Classes with no COCO equivalent constitute the COCO-disjoint set. This includes
goldfish, tiger shark, tree frog, ladybug, sulphur butterfly, wood rabbit,
hamster, wild boar, red fox, gibbon, ashcan, ballpoint, boathouse, chest,
clog, grand piano, iron, lab coat, padlock, table lamp, water tower, lemon,
carbonara, and agaric.

\section{AOL with pseudo-mask supervision.} To test whether AOL can be trained without ground-truth annotations, we evaluate a weakly supervised variant of Qwen2.5-VL in which AOL supervision is obtained from pseudo-masks generated by SAM3-I \cite{li2026sam3isegmentinstructions} on the RefCOCO training set. Given an image and its referring expression, SAM3-I, a dedicated segmentation model, generates a pseudo-mask for the target object. This pseudo-mask is used only during training to define positive and negative visual tokens for AOL. At inference time, the VLM is used normally, without SAM3-I or any segmentation model. As shown in Table~\ref{tab:pseudo_mask_aol}, pseudo-mask supervision improves cIoU from 66.2 to 69.6, approaching GT-mask supervision at 73.4. For reference, SAM3-I direct segmentation achieves 71.7 cIoU, while the pseudo-mask AOL model reaches 69.6 without a segmentation head.

\begin{table}[hbtp]
\centering
\small
\setlength{\tabcolsep}{4pt}
\renewcommand{\arraystretch}{1.08}
\begin{tabular}{@{}llc@{}}
\toprule
Model setting & AOL supervision & cIoU \\
\midrule
Pretrained Qwen2.5-VL & -- & 66.2 \\
NTP+AOL & SAM3-I pseudo-masks & 69.6 \\
NTP+AOL & Ground-truth masks & 73.4 \\
\bottomrule
\end{tabular}
\caption{RefCOCO validation cIoU after AOL training with SAM3-I pseudo-mask supervision on RefCOCO train. No frozen segmentation model is used at inference; masks are obtained by \textbf{Angular Relevance heatmaps + binarization}.}
\label{tab:pseudo_mask_aol}
\vspace{-3mm}
\end{table}

\section{Energy Pointing Game Evaluation}
\label{app:epg}

We additionally report Energy Pointing Game (EPG) as a complementary metric for evaluating localization quality. EPG measures the fraction of a heatmap's positive energy that falls inside the ground-truth mask, and therefore evaluates spatial alignment independently of any binarization threshold. As shown in Table~\ref{tab:imagenet_s50_ood_epg}, EPG scores consistently follow the same trend as mIoU.

\begin{table}[hbtp]
\centering
\small
\setlength{\tabcolsep}{3pt}
\renewcommand{\arraystretch}{1.08}
\begin{tabular}{@{}llcccccc@{}}
\toprule
Method & Setting
& \multicolumn{2}{c}{All}
& \multicolumn{2}{c}{Overlap}
& \multicolumn{2}{c}{Disjoint} \\
\cmidrule(lr){3-4}\cmidrule(lr){5-6}\cmidrule(lr){7-8}
& & mIoU & EPG & mIoU & EPG & mIoU & EPG \\
\midrule
Logit Lens
& Base model & 28.8 & 34.2 & 29.9 & 36.0 & 27.5 & 32.1 \\
\midrule
Ang. Rel.
& Base model & 36.9 & 39.0 & 39.8 & 39.7 & 33.5 & 38.2 \\
& NTP+AOL & \textbf{57.4} & \textbf{74.5} & \textbf{62.4} & \textbf{80.4} & \textbf{51.6} & \textbf{67.6} \\
\bottomrule
\end{tabular}
\caption{Full Localization results on ImageNet-S50.}
\label{tab:imagenet_s50_ood_epg}
\end{table}

\section{Referring Expression Comprehension task}

We also evaluate referring expression comprehension (REC) on the same datasets, which requires predicting a bounding box from the outputted text for the referred object. For REC, we use the same prompt as in RES that asks the VLM to output bounding box coordinates. Using identical prompts ensures that differences in performance reflect visual grounding quality rather than task-specific prompting.

We evaluate referring expression comprehension (REC) to verify that introducing AOL does not degrade the model’s ability to correctly identify the referred object while improving spatial grounding. In Table~\ref{tab:refcoco_dataset_performance}, models fine-tuned with NTP+AOL match or slightly improve upon both the base VLM and models fine-tuned with NTP alone across all datasets, indicating that enhanced visual grounding does not come at the cost of recognition accuracy. Also, our results are comparable to SOTA finetuned VLMs on REC, despite not introducing any task-specific grounding heads.

\begin{table}[hbtp]
\centering
\small
\setlength{\tabcolsep}{3pt}

\begin{tabular}{lccc}
\toprule
\textbf{Methods} & \textbf{RefCOCO} & \textbf{RefCOCO+} & \textbf{RefCOCOg} \\
\midrule

\multicolumn{4}{l}{\textit{LLaVA-v1.5-7B}} \\
Base Model & 52.90 & 42.55 & 45.00\\
NTP & 53.20 & 42.90 & 45.40\\
NTP + AOL & 54.15 & 43.60 &  46.20 \\
\midrule

\multicolumn{4}{l}{\textit{Qwen2.5-VL-7B}} \\
 Base Model & 90.05 & 85.40 & 86.60  \\
 NTP & 90.60 & 85.70 & 86.80 \\
NTP + AOL & \textbf{91.80} & \textbf{86.60} &  \textbf{87.50} \\
\midrule

% \multicolumn{4}{l}{\textit{SOTA (fine-tuning based VLMs)}} \\
% LLaVA-G-7B~{\tiny\cite{zhang2023llavagroundinggroundedvisualchat}} & 89.16 & 81.68 & 84.82 \\
% Shikra-7B~{\tiny\cite{chen2023shikraunleashingmultimodalllms}} & 87.0 & 81.6 & 82.3 \\
% Ferret-7B~{\tiny\cite{you2023ferretrefergroundgranularity}} & 87.5 & 80.8 & 83.9 \\
% Shikra-13B~{\tiny\cite{chen2023shikraunleashingmultimodalllms}} & 87.8 & 82.9 & 82.6 \\
% Ferret-13B~{\tiny\cite{you2023ferretrefergroundgranularity}} & 89.5 & 82.8 & 85.8 \\
% CogVLM-17B~{\tiny\cite{wang2024cogvlmvisualexpertpretrained}} &  \textcolor{red}{92.8} & \textcolor{red}{88.7} & \textcolor{red}{89.8} \\
\bottomrule
\end{tabular}

\caption{Comparison on the REC (Referring Expression Comprehension) task measured by IoU@0.5. 
% SOTA results correspond to fintuning–based VLMs.
}
\label{tab:refcoco_dataset_performance}
% \vspace{-4mm}
\end{table}

{
    \small
    \bibliographystyle{ieeenat_fullname}
    \bibliography{main}
}